\documentclass[11pt]{article}
\usepackage{acl}
\usepackage[utf8]{inputenc}
\usepackage[T1]{fontenc}
\usepackage{microtype}
\usepackage{inconsolata}
\usepackage{times}
\usepackage{latexsym}
\usepackage{attachfile}
\usepackage{csvsimple}
\usepackage{url}
\usepackage{amsmath,amssymb,amsthm,mathtools}
\usepackage{booktabs,multirow,array,tabularx}
\usepackage{algorithm}
\usepackage{algpseudocode}
\usepackage{float}
\usepackage{subcaption}
\usepackage{xcolor}
\usepackage{hyperref}
\usepackage{enumitem}
\usepackage{graphicx}
\hypersetup{colorlinks=true,citecolor=blue,linkcolor=blue,urlcolor=blue}

\newtheorem{proposition}{Proposition}

\usepackage{booktabs}
\usepackage{graphicx}   % for \resizebox
\usepackage{array}
\usepackage{pifont}     % for check/cross marks

\newcommand{\cmark}{\ding{51}}
\newcommand{\xmark}{\ding{55}}
\newcommand{\limitedmark}{\(\triangle\)}

\title{MENTIS: What Belief Changes Under Alignment? Measuring Multi-Scale Latent Torsion in Language Models}
\author{Anonymous}

\title{MENTIS: What Belief Changes Under Alignment? Measuring Multi-Scale Latent Torsion in Language Models}

\author{
  Partha Pratim Saha$^1$ \quad Samarth Raina$^2$ \quad Mayur Parvatikar$^1$ \quad Amit Dhanda$^{\dagger 4}$ \\
  Vinija Jain$^{\ddagger 5}$ \quad Aman Chadha$^{\S 6}$ \quad Amitava Das$^1$ \\[4pt]
  $^1$\textit{Pragya Lab, BITS Pilani Goa, India} \quad
  $^2$\textit{IIIT Delhi, India} \quad
  $^4$\textit{Amazon, USA} \\
  $^5$\textit{Meta, USA} \quad
  $^6$\textit{Apple, USA} \quad
  \setcounter{footnote}{1}\thanks{This work does not relate to Amit Dhanda's position at Amazon.}%
  \setcounter{footnote}{2}\thanks{This work does not relate to Vinija Jain's position at Meta.}%
  \setcounter{footnote}{3}\thanks{This work does not relate to Aman Chadha's position at Apple.}%
}

\begin{document}
\maketitle

\begin{abstract}

Preference alignment has substantially improved the observable behavior of large language models, yet it remains unclear what alignment changes \textbf{\emph{internally}}. Aligned systems still fail under jailbreaks, prompt injection, and retrieval-time corruption—suggesting \textbf{\emph{behavior-level evaluation alone is incomplete}}. Post-training should leave measurable traces in internal computation. We ask: when an IT model becomes a PA model, \textbf{\emph{what geometric structure changes}}, where do those changes concentrate, and how selectively do they vary across concepts, prompts, and model families?

We introduce \textbf{MENTIS}, a \emph{geometry-first} framework for measuring \textbf{\emph{alignment-induced internal reorganization}} in paired checkpoints. MENTIS compares IT and PA models using a primary layerwise covariance-based torsion norm, \textbf{T1}; a secondary spectral torsion diagnostic, \textbf{T2}; and an Energy-Radiance-Activation measure, \textbf{ERA}, for depth localization. Across four 7--8B model pairs on LITMUS, our study reveals that alignment-induced change is \textbf{\emph{selective rather than uniform}}: normative concepts exhibit larger torsion shifts than factual concepts on average; torsion is negatively correlated with contextual entropy; and peak effects localize to \textbf{\emph{architecture-specific mid-to-late layers}}. The same pattern appears across \textbf{(i)}\emph{word-level}, \textbf{(ii)}\emph{prompt-level}, and \textbf{(iii)}~\emph{model-level} analyses. These results suggest preference alignment leaves \textbf{\emph{structured, depth-localized geometric signatures}} in internal computation beyond what behavior-level evaluation alone can reveal. We release our \href{https://anonymous.4open.science/r/torsional-belief-vector-field/}{code, data}.

\end{abstract}

\subsection{What We Mean by Belief: A Task-Conditioned Directional View}
\label{sec:belief_definition}

The term \emph{belief} is easy to overinterpret in language-model analysis. We therefore use it in a deliberately operational sense. In this paper, a belief is not a symbolic proposition, a human-like mental state, or an independently recoverable variable inside the model. Instead, it denotes a \textbf{\emph{task-conditioned internal direction}}: a local direction in hidden-state space that increases support for a target continuation under a specified prompt, layer, and readout. Thus, when we speak of \emph{belief change}, we mean a measurable change in this internal directional organization, not a claim that the model has revised a human-style belief.

\begin{table*}[ht!]
\centering
\small
\setlength{\tabcolsep}{4.5pt}
\renewcommand{\arraystretch}{1.15}
\caption{\textbf{\emph{Positioning MENTIS relative to prior belief and alignment diagnostics.}}
Prior work studies behavioral outputs, preference-data values, factual associations, static hidden-state structure, or localized steering directions. MENTIS instead studies \textbf{\emph{paired-checkpoint geometric reorganization}}: how task-conditioned internal directions change from Instruction-Tuned (IT) to Preference-Aligned (PA) models across concepts, prompts, and depth.}
\label{tab:belief_positioning}
\resizebox{\linewidth}{!}{%
\begin{tabular}{lccccc}
\toprule
\textbf{Approach} &
\textbf{Behavior} &
\textbf{Belief-like} &
\textbf{Paired} &
\textbf{Depth} &
\textbf{Geometric} \\
&
\textbf{audit} &
\textbf{internals} &
\textbf{IT$\rightarrow$PA} &
\textbf{localized} &
\textbf{change} \\
\midrule
Output-level alignment evaluation~\citep{lee2024mechanistic,qi2024safety} &
\cmark & \xmark & \xmark & \xmark & \xmark \\

Preference-data value audits~\citep{obi2024valueimprint} &
\cmark & \xmark & \xmark & \xmark & \xmark \\

Factual-belief localization / editing~\citep{hase2023methods,meng2022locating} &
\xmark & \cmark & \xmark & \cmark & \xmark \\

Static residual-stream geometry~\citep{hosseini2023straighten,shai2024beliefgeometry} &
\xmark & \cmark & \xmark & \cmark & \xmark \\

Representation engineering / steering directions~\citep{zou2023repe,arditi2024refusal} &
\xmark & \cmark & \xmark & \cmark & \xmark \\

Checkpoint drift / CKA-style similarity~\citep{kornblith2019similarity} &
\xmark & \xmark & \cmark & \limitedmark & \limitedmark \\

\hline

\textbf{MENTIS} &
\xmark & \cmark & \cmark & \cmark & \cmark \\
\bottomrule
\end{tabular}}
\vspace{0.25em}

\footnotesize{
\textbf{Legend:} \cmark = explicitly supported; \xmark = not the primary object of study; \limitedmark = partially supported but without task-conditioned directional torsion.
}
\end{table*}

This distinction is central to our framing. Prior work has shown that language models contain measurable internal structure: factual associations can be localized or edited~\citep{hase2023methods,meng2022locating}, residual streams can encode belief-state-like geometry~\citep{hosseini2023straighten,shai2024beliefgeometry}, and safety or refusal behavior can sometimes be mediated by low-dimensional activation directions~\citep{zou2023repe,arditi2024refusal}. Other work studies alignment behavior or value structure through outputs and preference data~\citep{lee2024mechanistic,qi2024safety,obi2024valueimprint}. MENTIS asks a different question: \textbf{\emph{when an instruction-tuned checkpoint is transformed into a preference-aligned checkpoint, how does the task-conditioned internal direction itself reorganize across depth?}}

% This paired-checkpoint view is important because a static analysis of one aligned model cannot isolate what alignment changed. 
A single-model probe may reveal that a representation is linearly decodable, or that a factual association is localized, but it does not directly measure the geometric update induced by preference alignment. MENTIS therefore treats belief change as \textbf{\emph{alignment-induced directional reorientation}}: we compare IT and PA checkpoints under matched prompts and matched target continuations, then measure how the associated internal directions change in magnitude, angle, spectral structure, and depth localization.

Table~\ref{tab:belief_positioning} summarizes this positioning. The key distinction is that MENTIS is not primarily a behavioral audit, a preference-data audit, or a static probe of hidden states. It is a \textbf{\emph{comparative geometric diagnostic}} for measuring how alignment changes task-conditioned internal organization across words, prompts, and model families.

\section{MENTIS: Alignment as Internal Geometric Reorganization}
\label{sec:method}

\subsection{Operational View: Belief as Directional Support}

MENTIS treats preference alignment as a \textbf{\emph{geometric transformation}} of internal computation. If post-training changes how a model supports, suppresses, or redirects task-relevant continuations, then that change should leave measurable traces in how hidden-state directions evolve across depth.

We use ``belief'' only in an operational sense. It does not denote a symbolic proposition or a human-like mental state. Instead, it refers to a \textbf{\emph{task-conditioned internal direction}}: the local direction in representation space that increases support for a target continuation. Thus, \emph{belief change} means alignment-induced reorientation of this directional field, not philosophical belief revision.

This gives MENTIS a narrow goal: compare an Instruction-Tuned (IT) checkpoint and its Preference-Aligned (PA) counterpart under matched prompts and targets, then ask where the internal geometry changes, how strongly it changes, and whether the change is selective across concepts, prompts, and model families.

\subsection{Paired-Checkpoint Setup}

Let \(\theta^{(0)}\) denote the IT model and \(\theta^{(1)}\) its PA counterpart. For prompt \(x\), let
\[
h_{\ell}^{(m)}(x)\in\mathbb{R}^{d},
\qquad
m\in\{0,1\},\quad \ell\in\{1,\dots,L\},
\]
be the hidden representation at layer \(\ell\). Each prompt induces a depth-indexed trajectory
\[
\mathcal{H}^{(m)}(x)
=
\bigl(
h_{1}^{(m)}(x),
h_{2}^{(m)}(x),
\dots,
h_{L}^{(m)}(x)
\bigr).
\]
We also consider semantic subsets \(\mathcal{P}^{(c)}\) for value category \(c\) and \(\mathcal{P}^{(w)}\) for prompts associated with concept \(w\).

\subsection{Layerwise Directional Field}

Using the unembedding or equivalent readout operator \(W_U\), we define a layer-indexed predictive distribution:
\[
p_{\theta^{(m)},\ell}(y\mid x)
=
\operatorname{softmax}\!\left(W_U h_{\ell}^{(m)}(x)\right)_y .
\]
In experiments, \(y\) is the first token of the gold response for each LITMUS~\cite{Litmas_AQI} prompt. This keeps the readout tied to a concrete predictive object rather than to a pooled sequence-level surrogate.

The layerwise directional field is
\[
v_{\ell}^{(m)}(x)
:=
\nabla_{h_{\ell}^{(m)}(x)}
\log p_{\theta^{(m)},\ell}(y\mid x).
\]
Geometrically, \(v_{\ell}^{(m)}(x)\) is the steepest local direction that increases support for \(y\) at layer \(\ell\). For a semantic subset \(\mathcal{Q}\subseteq\mathcal{P}\), we define
\[
v_{\ell}^{(m,\mathcal{Q})}
:=
\mathbb{E}_{x\sim\mathcal{Q}}
\left[
v_{\ell}^{(m)}(x)
\right].
\]

\begin{figure*}[ht!]
\centering

\begin{minipage}[t][0.46\textheight][c]{0.49\textwidth}
\centering
\vfill
\includegraphics[width=\linewidth]{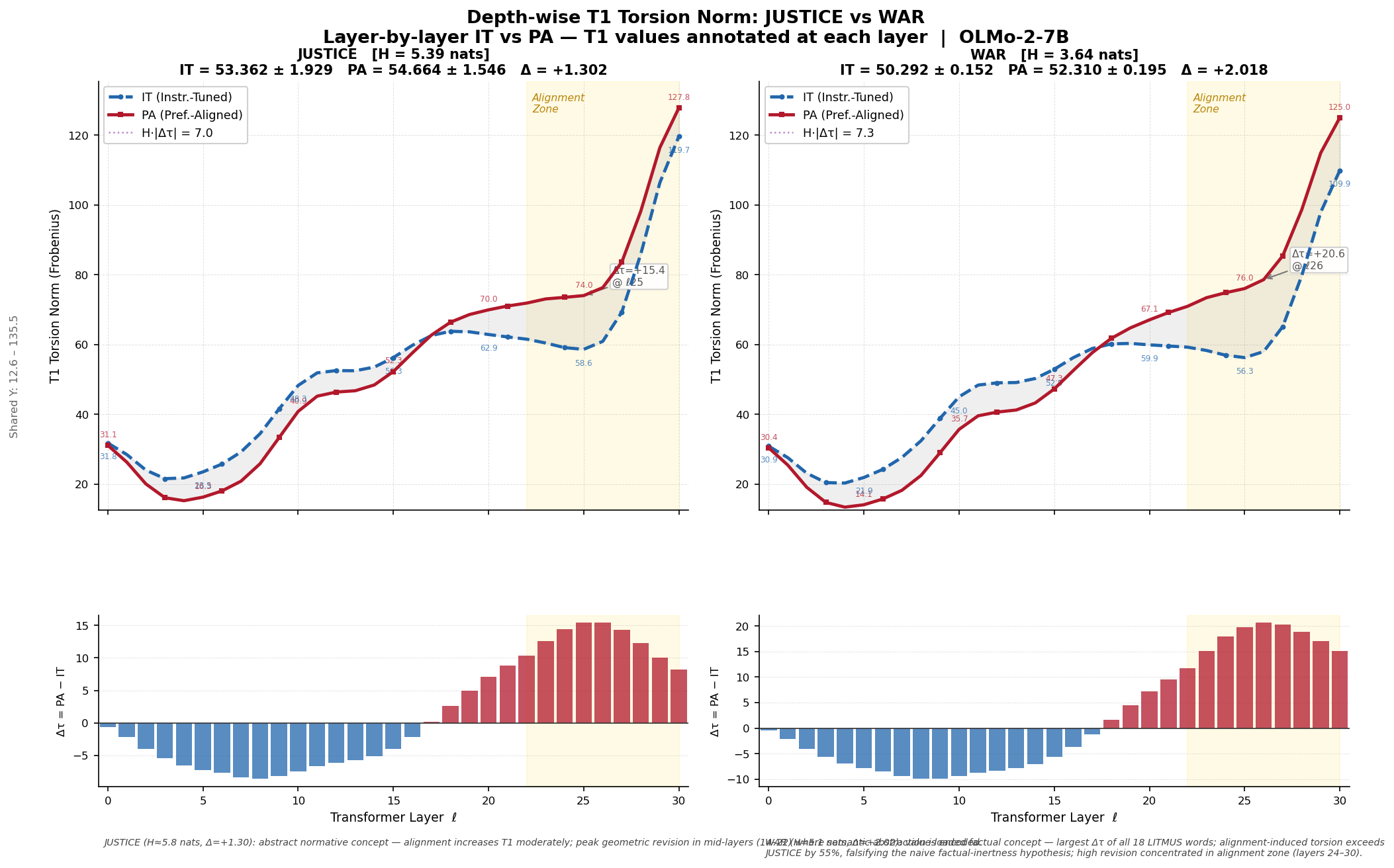}
\vfill
\caption{
\textbf{\emph{Depthwise torsion separates value-salient concepts.}}
Layerwise T1 torsion profiles for \textbf{JUSTICE} and \textbf{WAR} under Instruction-Tuned (IT) and Preference-Aligned (PA) checkpoints. Alignment-induced reorganization is not uniform across concepts: the two semantic regions exhibit distinct torsional trajectories and different late-layer concentration patterns.
}
\label{fig:justice_vs_war}
\end{minipage}
\hfill
\begin{minipage}[t][0.46\textheight][c]{0.49\textwidth}
\centering
\includegraphics[width=0.85\linewidth]{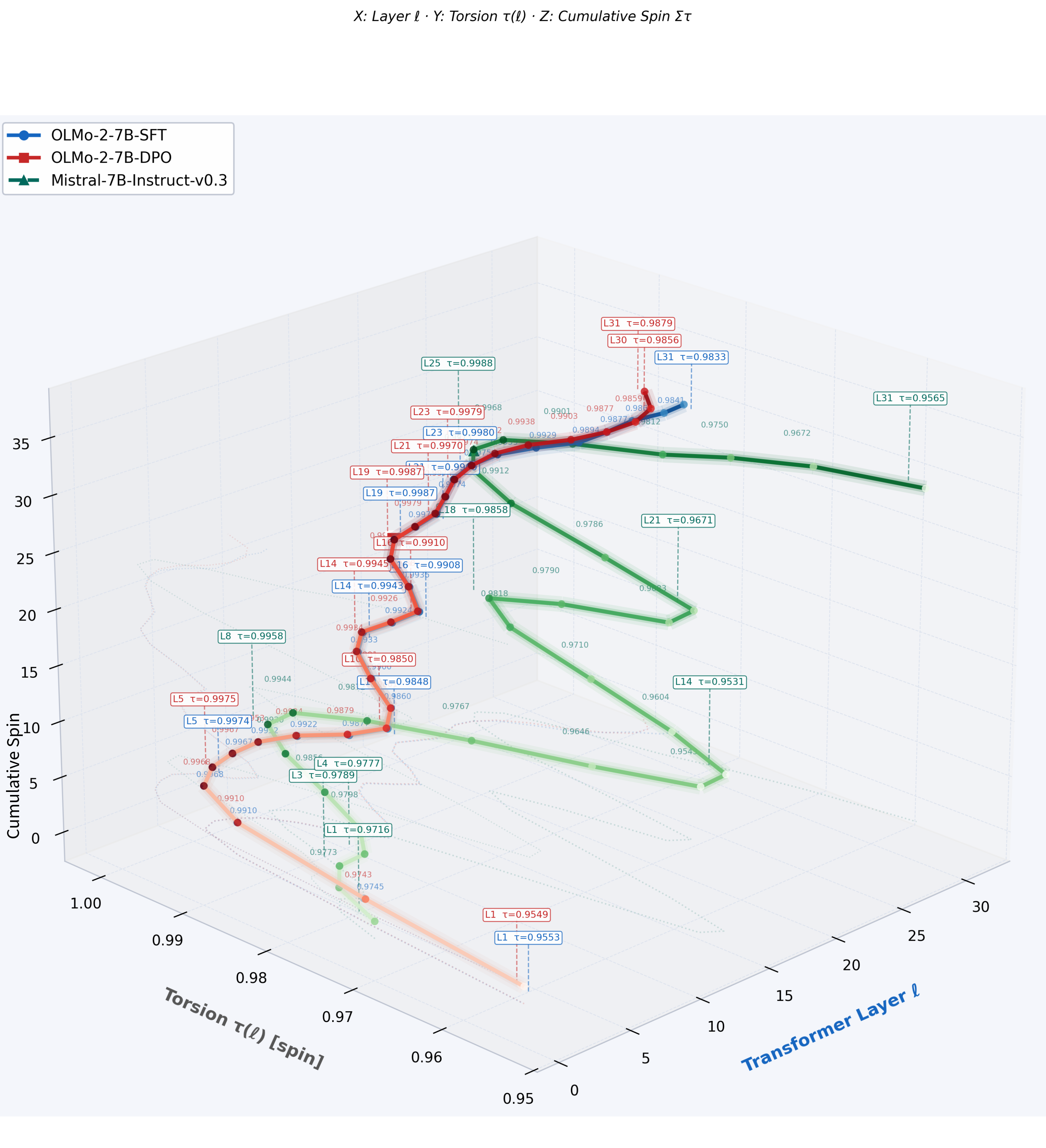}
\vspace{-0.5em}
\caption{
\textbf{\emph{Torsion geometry of JUSTICE across depth.}}
Three-dimensional depthwise torsion trajectory for \textbf{JUSTICE} under IT and PA checkpoints, visualizing how alignment reshapes the concept-specific representational path across layers.
}
\label{fig:torsion_3d_justice}
\end{minipage}

\vspace{-0.7em}
\end{figure*}

Figure~\ref{fig:justice_vs_war} provides two motivating examples of this depthwise reorganization, showing that alignment changes torsion profiles differently across concept pairs.

\subsection{Alignment-Induced Directional Shift}

The local IT\(\rightarrow\)PA displacement at layer \(\ell\) is
\[
\Delta v_{\ell}(x)
=
v_{\ell}^{(1)}(x)-v_{\ell}^{(0)}(x).
\]
We summarize this displacement through magnitude and angle:
\[
\Delta_{\mathrm{mag},\ell}(x)
=
\left|
\|v_{\ell}^{(1)}(x)\|-
\|v_{\ell}^{(0)}(x)\|
\right|,
\]
\[
\Delta_{\mathrm{ang},\ell}(x)
=
\arccos
\left(
\frac{
\langle v_{\ell}^{(0)}(x),v_{\ell}^{(1)}(x)\rangle
}{
\|v_{\ell}^{(0)}(x)\|\,
\|v_{\ell}^{(1)}(x)\|
}
\right).
\]
Magnitude asks whether alignment changes the strength of the local support direction; angle asks whether it rotates the direction itself.

\subsection{Depthwise Torsion}

A model may preserve local support at one layer while changing how that support evolves across depth. We therefore measure cumulative directional turning. For model \(m\), the adjacent-layer turning angle is
\[
\omega_{\ell}^{(m)}(x)
=
\arccos
\left(
\frac{
\langle v_{\ell}^{(m)}(x),v_{\ell+1}^{(m)}(x)\rangle
}{
\|v_{\ell}^{(m)}(x)\|\,
\|v_{\ell+1}^{(m)}(x)\|
}
\right).
\]
The total angular torsion is
\[
\mathcal{T}^{(m)}(x)
=
\sum_{\ell=1}^{L-1}
\omega_{\ell}^{(m)}(x).
\]
The alignment-induced torsional shift is
\[
\Delta\mathcal{T}(x)
=
\mathcal{T}^{(1)}(x)-\mathcal{T}^{(0)}(x).
\]
A positive value indicates that the PA model exhibits greater cumulative directional turning than its IT counterpart.

\begin{proposition}[Non-negativity and zero condition]
For any model \(m\) and prompt \(x\), \(\mathcal{T}^{(m)}(x)\geq 0\). Moreover, \(\mathcal{T}^{(m)}(x)=0\) if and only if every adjacent pair \(v_{\ell}^{(m)}(x),v_{\ell+1}^{(m)}(x)\) is co-directional.
\end{proposition}

\begin{figure*}[ht!]
\centering
\includegraphics[width=\linewidth]{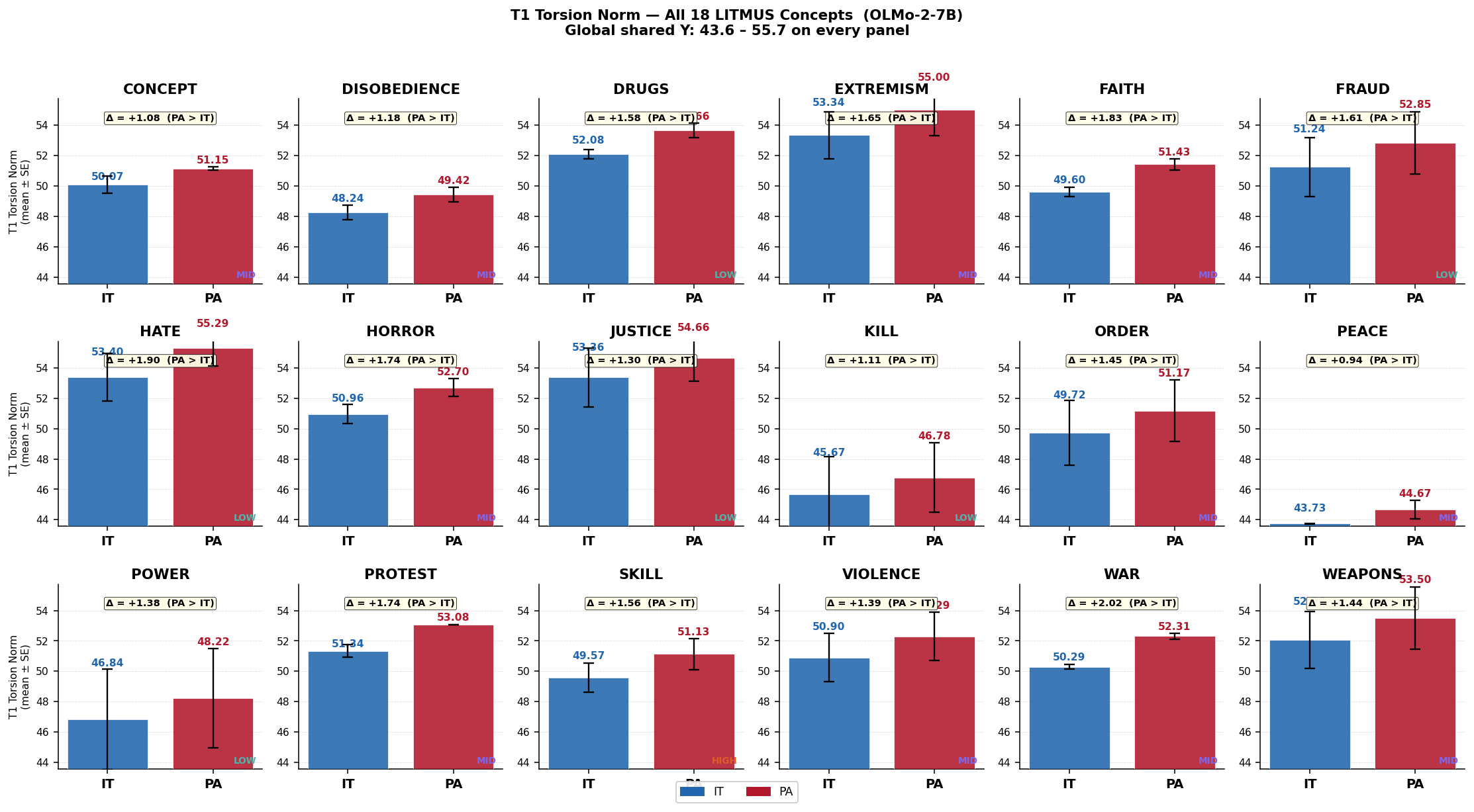}
\caption{
\textbf{\emph{Concept-level torsion is selective rather than uniform.}}
Shared-axis torsion summaries for 18 LITMUS concepts. The figure shows that alignment-induced geometric change varies substantially across concepts, supporting the claim that preference alignment reorganizes internal directions selectively rather than applying a uniform representational shift.
}
\label{fig:18_words}
\end{figure*}

\subsection{Matrix Torsion: T1, T2, and ERA}

Angular torsion gives a prompt-level directional summary. To obtain a layerwise matrix-level view, we define torsion from token-by-hidden-state matrices. For \(H_\ell^{(m)}(x)\in\mathbb{R}^{T\times d}\), let \(C_\ell^{(m)}\) be the token-centered hidden-state matrix. We define the cross-layer covariance
\[
M_\ell^{(m)}
=
\frac{1}{T}
\bigl(C_\ell^{(m)}\bigr)^\top
C_{\ell+1}^{(m)}.
\]
Its skew-symmetric component is
\[
S_\ell^{(m)}
=
\frac{
M_\ell^{(m)}-\bigl(M_\ell^{(m)}\bigr)^\top
}{2}.
\]

We use two complementary torsion diagnostics:
\[
\tau_{\mathrm{T1},\ell}^{(m)}
=
\|S_\ell^{(m)}\|_F,
\]
\[
\tau_{\mathrm{T2},\ell}^{(m)}
=
\mathrm{Var}
\left(
\left\{
|\Im\lambda_j(S_\ell^{(m)})|:\lambda_j\neq 0
\right\}
\right).
\]

\textbf{T1} is the primary metric: it measures total rotational energy at layer \(\ell\). \textbf{T2} is a secondary spectral diagnostic: it measures how unevenly that rotational signal is distributed across modes. The skew-symmetric component is central because it isolates rotational layer-to-layer change, whereas the symmetric component mixes co-variation with scale.

\begin{proposition}[T1 as a matrix-level turning proxy]
Let \(S_\ell^{(m)}\) be the skew-symmetric cross-layer component and let \(\sigma_1(S_\ell^{(m)})\) be its largest singular value. Larger \(\sigma_1(S_\ell^{(m)})\) implies a larger principal rotational component of layer-to-layer change. Thus, T1 provides a matrix-level proxy for the same turning phenomenon summarized by angular torsion.
\end{proposition}

Finally, \textbf{Energy-Radiance-Activation} (ERA) localizes where alignment-induced torsion concentrates. Let
\[
\Delta\tau_{\ell}
=
\tau_{\mathrm{T1},\ell}^{(1)}
-
\tau_{\mathrm{T1},\ell}^{(0)}.
\]
ERA normalizes the layerwise difference profile and selects the peak layer:
\[
\ell^\star
=
\arg\max_{\ell}
\operatorname{Norm}
\left(
\Delta\tau_{\ell}
\right).
\]
The resulting \(\ell^\star\) gives a depth address for later patching, steering, or causal intervention.

\subsection{Multi-Scale Measurement}

MENTIS operates at three linked scales.

\begin{itemize}[leftmargin=1.2em]
    \item \textbf{Word level:} concept-conditioned torsion shifts \(\Delta\mathcal{T}(w)\), T1, and T2 reveal which concepts undergo the largest alignment-induced reorientation.

    \item \textbf{Prompt level:} prompt-wise torsion shifts \(\Delta\mathcal{T}(x)\) quantify how safe and unsafe prompt families differ in their internal geometric response.

    \item \textbf{Model level:} layerwise profiles and band-concentration ratios compare whether alignment effects are shallow, mid-layer, late-layer, or distributed across architectures.
\end{itemize}

Figure~\ref{fig:18_words} summarizes concept-level selectivity across the 18 representative LITMUS concepts.

\subsection{Practical Estimation}

We project hidden states to 64 PCA dimensions and use \(\epsilon\)-clipped cosine similarity for numerical stability. Same-model null comparisons produce torsion shifts below \(10^{-4}\), suggesting that measured IT\(\rightarrow\)PA differences are not dominated by numerical noise.

These choices serve two purposes. First, PCA provides a shared comparison space across model families with different hidden dimensions. Second, clipped cosine similarity prevents small-norm numerical artifacts from dominating angular estimates. These controls do not by themselves prove robustness, but they reduce the risk that the reported torsion effects arise from raw dimensionality, scale, or floating-point instability.

\section{Experimental Setup}
\label{sec:setup}

\paragraph{Models.}
We evaluate four paired Instruction-Tuned to Preference-Aligned (IT$\rightarrow$PA) model families: OLMo-2-1124-7B~\citep{teamolmo2025olmo2}, Mistral-7B-v0.3~\citep{jiang2023mistral}, Llama-3.1-8B~\citep{grattafiori2024llama3}, and T\"ulu-3-8B~\citep{lambert2024tulu3}. We focus on 7--8B decoder-only models for three reasons. First, paired IT and PA checkpoints are publicly available at this scale. Second, these models provide sufficient layer resolution for depth-localization analyses. Third, keeping parameter count roughly fixed reduces the risk that cross-model torsion differences merely reflect scale rather than post-training dynamics.

\paragraph{Benchmark.}
We use LITMUS~\citep{Litmas_AQI}, a concept-structured alignment benchmark with 20,439 prompts, 7 value axioms, 18 representative concepts, 22,631 content words, and safety-agreement score \(\kappa=0.82\). LITMUS is well suited to MENTIS because our goal is not only to detect whether IT and PA checkpoints differ, but to characterize \textbf{\emph{where}} that difference concentrates across concepts, prompt families, and depth.

This requires three benchmark properties: structured semantic partitions, explicit safe--unsafe labels, and enough concept-level coverage for word-level torsion analysis. LITMUS provides this combination directly. We therefore treat it as a \textbf{\emph{measurement testbed}} for internal alignment geometry, not as a claim of full benchmark coverage.

Figure~\ref{fig:justice_vs_war} serves as concept-pair anchors for the broader 18-concept analysis. \textsc{PEACE} and \textsc{PROTEST} occupy a related value region but differ in normative contestation, making them useful for testing within-domain selectivity. \textsc{JUSTICE} and \textsc{WAR} cut across the normative--factual boundary, testing whether torsion merely reproduces the benchmark taxonomy or captures alignment-salient geometric change.

\paragraph{Baselines.}
We compare MENTIS with three complementary baselines. CKA~\citep{kornblith2019similarity} measures representational similarity across checkpoints. Cosine distance between paired hidden-state summaries captures checkpoint drift without directional structure. RepE~\citep{zou2023repe} provides a linear-direction baseline for representation-level steering. Together, these baselines cover similarity, distance, and linear-subspace views. MENTIS is useful if torsion adds concept discrimination or depth localization beyond these alternatives.

\paragraph{Statistical protocol.}
The prompt is the unit of analysis for prompt-level tests; the concept is the unit of analysis for concept-level summaries unless otherwise stated. We use Mann--Whitney \(U\) tests with Holm--Bonferroni correction for four-model comparisons, Cohen's \(d\) for effect size, and 95\% bootstrap confidence intervals with 2,000 resamples. We report concept-level summaries for semantic-selectivity claims and prompt-level summaries for safe--unsafe distributional comparisons.

\section{Empirical Structure of IT$\rightarrow$PA Reorganization}
\label{sec:results}

We evaluate MENTIS as an operational diagnostic of \textbf{\emph{alignment-induced internal reorganization}}. The analysis focuses on five questions. (i) Does torsion separate IT$\rightarrow$PA change from generic representation drift? (ii) Is the change uniform across semantic content, or concentrated in specific concept regions? (iii) How does torsion relate to contextual entropy? (iv) Does reorganization localize to architecture-specific depth bands? (v) Do safe and unsafe prompts induce different geometric responses after alignment?

\begin{figure*}[t]
\centering

\begin{minipage}[t]{0.49\textwidth}
\centering
\includegraphics[width=\linewidth,height=4cm]{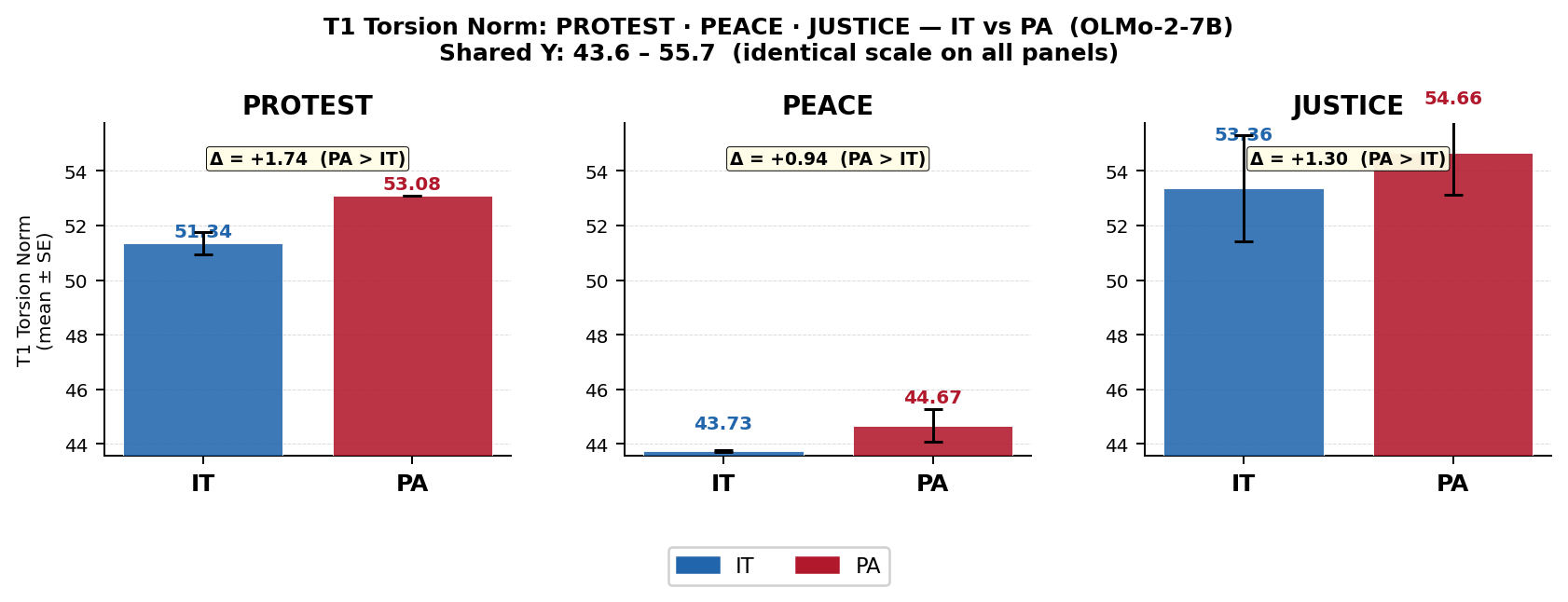}
\vspace{-0.5em}
\caption{
\textbf{\emph{Alignment-induced torsion is semantically selective.}}
In OLMo-2-1124-7B, \textbf{JUSTICE}, \textbf{PEACE}, and \textbf{PROTEST} exhibit different IT$\rightarrow$PA torsion increases. Alignment-induced reorganization therefore varies within value-related concepts rather than appearing as a uniform semantic offset.
}
\label{fig:protest_peace}
\end{minipage}
\hfill
\begin{minipage}[t]{0.49\textwidth}
\centering
\includegraphics[width=\linewidth,height=4cm]{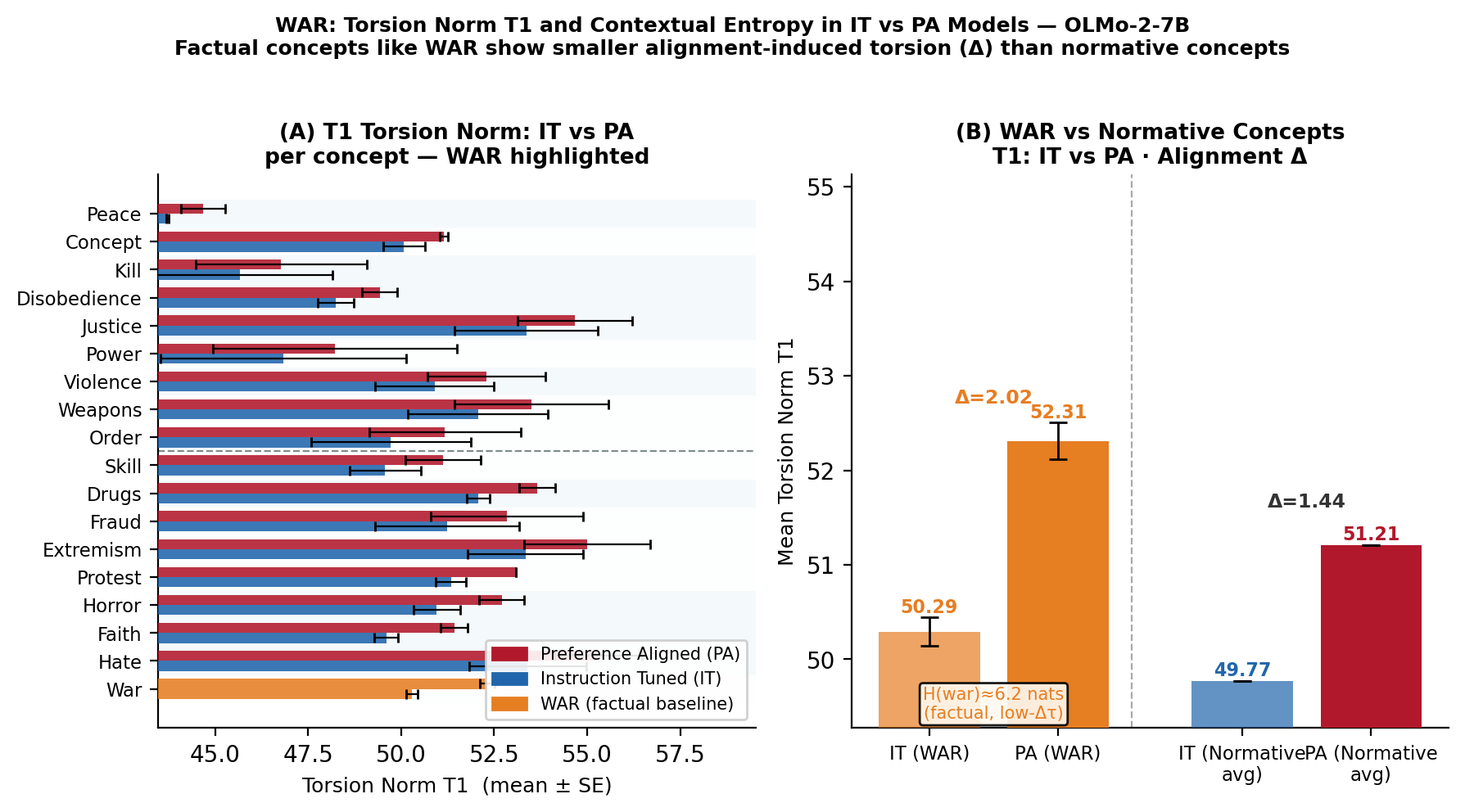}
\vspace{-0.5em}
\caption{
\textbf{\emph{Torsion decreases with contextual entropy.}}
For \textbf{WAR}, higher contextual entropy is associated with a lower torsion norm in both IT and PA checkpoints. The strongest alignment-induced reorganizations, therefore, appear in semantically structured regions rather than in maximally uncertain contexts.
}
\label{fig:Torsion_vs_entropy}
\end{minipage}

\vspace{-0.7em}
\end{figure*}

\subsection{Torsion Separates Alignment Change from Generic Drift}

A useful alignment diagnostic should do more than detect that two checkpoints differ. It should identify \textbf{\emph{how}} they differ: whether the change is directional rather than purely metric, whether it varies across concepts, and whether it concentrates at specific layers. Table~\ref{tab:baseline} compares MENTIS against three standard alternatives: CKA, cosine distance, and RepE.

MENTIS produces strongest concept-level discrimination (coefficient of Variation)
\[
\mathrm{CV}_{\mathrm{MENTIS}}=0.64,
\]
compared with
\vspace{-5pt}
\[
\mathrm{CV}_{\mathrm{CKA}}=0.08,
\qquad
\mathrm{CV}_{\mathrm{RepE}}=0.21.
\]
Thus, torsion varies across concepts, global similarity and linear-direction summaries remain comparatively compressed. It measures variation in
\vspace{-5pt}
\[
\Delta \tau(c)
=
\mathbb{E}_{x\in \mathcal{P}^{(c)}}
\left[
\tau^{(1)}(x)-\tau^{(0)}(x)
\right],
\]
\vspace{-3pt}
rather than only absolute displacement between hidden states. Two checkpoints can be far apart in Euclidean representation space while preserving nearly the same local support direction.
For safe–unsafe discrimination, MENTIS reaches,
\vspace{-5pt}
\[
\mathrm{AUC}=0.89,
\qquad
95\%~\mathrm{CI}=[0.85,0.93],
\]
while cosine distance reaches only \(0.64\). Same-model IT$\rightarrow$IT null comparisons produce results
\[
\Delta\tau<10^{-4},
\]
indicating that the measured IT$\rightarrow$PA torsion shifts are not dominated by numerical instability or trivial checkpoint variation.

The comparison also exposes the structural limitations of the baselines. CKA compresses a rich layerwise trajectory into a single similarity score. Cosine distance measures displacement without distinguishing directional rotation from ordinary drift. RepE captures linear activation directions but does not model cumulative turning across depth. MENTIS adds precisely this missing structure: it measures how \textbf{\emph{task-conditioned internal directions rotate across layers}} and whether that rotation is semantically selective. In particular, MENTIS compares the depthwise path
\vspace{-5pt}
\[
v_1(x)\rightarrow v_2(x)\rightarrow \cdots \rightarrow v_L(x)
\]

rather than only endpoint representations. The relevant object is therefore the torsional shift
\[
\Delta \mathcal{T}(x)
=
\sum_{\ell=1}^{L-1}
\left(
\omega_{\ell}^{(1)}(x)-\omega_{\ell}^{(0)}(x)
\right),
\]
which separates \textbf{\emph{directional reorientation}} from generic checkpoint drift.

\begin{table}[t]
\centering
\small
\caption{
\textbf{\emph{MENTIS separates structured alignment change from generic drift.}}
Concept CV measures variation across representative concepts; AUC measures safe--unsafe discrimination; Depth indicates whether the method provides an intrinsic localization signal comparable to ERA. The same-model null verifies that the observed IT$\rightarrow$PA shifts are not explained by numerical noise alone.
}

\label{tab:baseline}
\begin{tabular}{lccc}
\toprule
Method & Concept CV & AUC & Depth \\
\midrule
CKA & 0.08 & -- & No \\
Cosine distance & -- & 0.64 & Limited \\
RepE & 0.21 & -- & No \\
\textbf{MENTIS} & \textbf{0.64} & \textbf{0.89} & \textbf{Yes} \\
Same-model null & \(<10^{-4}\) & -- & -- \\
\bottomrule
\end{tabular}
\end{table}

The operational conclusion is stronger than simple checkpoint separability. IT$\rightarrow$PA updates exhibit \textbf{\emph{measurable torsional structure}}, as shown in Fig.~\ref{fig:torsion_cartography_B}, that is more discriminative across concepts and more localizable across depth than the baseline summaries tested here.
\vspace{-5pt}
\subsection{Alignment-Induced Reorganization is Semantically Selective}

We next ask whether alignment-induced reorganization is uniform across semantic content. If preference alignment acted as a broad global offset, then most concepts would exhibit similar torsion shifts. Instead, the observed changes are \textbf{\emph{selective}}.

At the category level, normative concepts show larger T1 shifts than factual concepts:
\[
\Delta \mathrm{T1}_{\mathrm{normative}}=1.74,
\qquad
\Delta \mathrm{T1}_{\mathrm{factual}}=1.47.
\]
The separation is statistically large relative to within-category variance:
\[
d=1.8,
\\
95\%~\mathrm{CI}=[1.2,2.4],
\\
p_{\mathrm{perm}}<0.001.
\]
Equivalently, if
\begin{align*}
\mu_{\mathrm{norm}}
&=
\mathbb{E}_{c\in\mathcal{C}_{\mathrm{norm}}}
\left[
\Delta \mathrm{T1}(c)
\right],
\\
\mu_{\mathrm{fact}}
&=
\mathbb{E}_{c\in\mathcal{C}_{\mathrm{fact}}}
\left[
\Delta \mathrm{T1}(c)
\right].
\end{align*}
the observed effect satisfies

\[
\mu_{\mathrm{norm}}-\mu_{\mathrm{fact}}>0.
\]
Thus,
\[
\Delta \mathrm{T1}_{\mathrm{normative}}
>
\Delta \mathrm{T1}_{\mathrm{factual}},
\]
supporting the view that alignment-induced internal change concentrates in \textbf{\emph{value-salient semantic regions}}.

The selectivity also appears within related concept groups. Figure~\ref{fig:protest_peace} compares IT$\rightarrow$PA torsion shifts for \textsc{JUSTICE}, \textsc{PEACE}, and \textsc{PROTEST} in OLMo-2-1124-7B. Although these concepts occupy related normative regions, they do not exhibit the same torsion response. MENTIS is therefore not simply recovering a coarse semantic partition. It measures \textbf{\emph{finer-grained directional reorganization}} within a semantic neighborhood.

The effect is not reducible to the benchmark's normative-factual split: although \textsc{WAR} is labelled factual, it exhibits the largest single-concept torsion shift, suggesting that MENTIS tracks \textbf{\emph{alignment-salient framing}} rather than merely reproducing dataset labels. MENTIS estimates a continuous score,
\vspace{-5pt}
\[
c \mapsto \Delta \mathrm{T1}(c),
\]
not a binary recovery of the benchmark label.

Figure~\ref{fig:18_words} provides the broader concept-level profile. The 18-concept distribution confirms that torsion is unevenly distributed across concepts: some concepts undergo strong IT$\rightarrow$PA reorganization, while others remain comparatively stable.

\subsection{Torsion Concentrates in Structured, Lower-Entropy Contexts}

We next examine whether torsion is related to contextual uncertainty. Let \(H(x)\) denote contextual entropy and \(\tau_{\mathrm{T1}}(x)\) the corresponding torsion norm. If torsion merely tracked uncertainty, one would expect larger shifts in high-entropy regions. The observed relationship is the opposite:
\vspace{-5pt}
\[
\rho\!\left(\tau_{\mathrm{T1}},H\right)<0.
\]
\vspace{-2pt}
Figure~\ref{fig:Torsion_vs_entropy} shows this relationship for \textsc{WAR}. In OLMo, the Spearman correlation is,
\vspace{-5pt}
\[
\rho=-0.164,
\qquad
p=3.04\times10^{-6},
\]
while in Mistral, the effect is stronger.
\vspace{-5pt}
\[
\rho=-0.387,
\qquad
p=5.43\times10^{-30}.
\]

The negative relationship is therefore statistically reliable across both architectures.

\vspace{-2pt}
Operationally, this suggests that alignment does not rotate internal directions uniformly wherever the model is uncertain. Instead, torsion is strongest where the context is structured enough to induce stable task-conditioned directions yet sufficiently alignment-salient for post-training to modify them. In mathematical terms, the result supports a regime of
\vspace{-4pt}
\[
\mathbb{E}[\tau_{\mathrm{T1}}\mid H \in \mathrm{low/mid}]
>
\mathbb{E}[\tau_{\mathrm{T1}}\mid H \in \mathrm{high}],
\]
rather than monotone uncertainty amplification. This entropy pattern is therefore consistent with the semantic-selectivity result above.

Importantly, the result should not be interpreted causally. MENTIS does not establish that entropy controls torsion. The contribution is diagnostic: the observed pattern rules out the simplest explanation that torsion is merely a byproduct of \textbf{\emph{generic uncertainty}}.

\subsection{ERA Identifies Architecture-Specific Depth Addresses}

MENTIS is designed not only to measure whether IT$\rightarrow$PA reorganization occurs, but also to localize where it concentrates. ERA summarizes the layerwise torsion-difference profile and identifies the peak depth
\vspace{-5pt}
\[
\ell^\star
=
\arg\max_{\ell}
\operatorname{Norm}
\!\left(
\tau_{\mathrm{T1},\ell}^{(1)}
-
\tau_{\mathrm{T1},\ell}^{(0)}
\right).
\]

Table~\ref{tab:depth} reports the resulting peak layers. The peaks are architecture-specific. OLMo-2-1124-7B and T\"ulu-3-8B peak late, at layers \(29\)--\(30\). Llama-3.1-8B peaks at layer \(20\), while Mistral-7B-v0.3 peaks earlier, around layer \(14\).

\begin{table}[t]
\centering
\small
\caption{
\textbf{\emph{ERA localizes alignment-induced reorganization to architecture-specific depth bands.}}
Peak layer \(\ell^\star\) denotes the layer band where the IT$\rightarrow$PA torsion difference is largest. Different architectures peak at different depths, indicating that alignment-induced geometric change is model-family specific rather than fixed at a universal layer.
}

\label{tab:depth}
\begin{tabular}{lc}
\toprule
Model pair & Peak layer \(\ell^\star\) \\
\midrule
OLMo-2-1124-7B & 29--30 \\
T\"ulu-3-8B & 29--30 \\
Llama-3.1-8B & 20 \\
Mistral-7B-v0.3 & 14 \\
\bottomrule
\end{tabular}
\end{table}

Thus, alignment-induced torsion does not emerge at a universal transformer depth. Instead, the IT$\rightarrow$PA update interacts with the \textbf{\emph{representational organization of each architecture}}. We can view the ERA profile as a normalized layerwise density,
\vspace{-10pt}
\[
r_{\ell}
=
\frac{
|\tau_{\mathrm{T1},\ell}^{(1)}-\tau_{\mathrm{T1},\ell}^{(0)}|
}{
\sum_{j=1}^{L}
|\tau_{\mathrm{T1},j}^{(1)}-\tau_{\mathrm{T1},j}^{(0)}|
},
\]
\vspace{-4pt}
so that \(\ell^\star=\arg\max_\ell r_\ell\). This makes depth localization comparable across architectures even when raw torsion magnitudes differ.

% ERA is methodologically important. A metric that only says two checkpoints differ provides no direct target for mechanistic follow-up. 
ERA converts a descriptive geometric profile into a \textbf{\emph{depth address}}. It identifies where patching, steering, or ablation should begin. The result does not establish that the peak layer is causally necessary for aligned behavior, but it makes subsequent causal analysis more constrained and testable.

Figure~\ref{fig:justice_vs_war} provides qualitative anchors for this aggregate depth result. In both concept-pair profiles, the IT--PA separation is concentrated in specific layer bands rather than distributed uniformly across the entire network.

This finding is also broadly consistent with prior evidence that alignment effects can be localized and architecture-dependent~\citep{lee2024mechanistic,qi2024safety}. MENTIS adds a complementary measurement perspective: rather than localizing a refusal direction or behavioral trigger, it localizes the peak of \textbf{\emph{torsional reorganization}} in the task-conditioned directional field.

\subsection{Safe Prompts Reveal Reorganization Beyond Refusal}

A shallow-refusal account would predict the largest internal changes on unsafe prompts, but in our benchmark setting we observe the opposite pattern: safe prompts induce larger torsional shifts across all four architectures.
\[
\mathbb{E}[\Delta\tau\mid \mathrm{safe}]
>
\mathbb{E}[\Delta\tau\mid \mathrm{unsafe}],
\]
with corrected
\[
p<10^{-19}.
\]

We interpret this result conservatively. It does not prove that safe prompts are causally central to alignment. It establishes that, in this benchmark, preference alignment modifies the internal directional geometry of safe prompts more strongly than that of unsafe prompts. More explicitly, if \(\mathcal{P}_{s}\) and \(\mathcal{P}_{u}\) denote safe and unsafe prompt sets, the observed contrast is
\[
\frac{1}{|\mathcal{P}_{s}|}
\sum_{x\in \mathcal{P}_{s}}
\Delta\tau(x)
>
\frac{1}{|\mathcal{P}_{u}|}
\sum_{x\in \mathcal{P}_{u}}
\Delta\tau(x).
\]

This outcome moves the analysis beyond refusal-only explanations. Unsafe prompts may activate comparatively stereotyped refusal pathways, producing less variation in the task-conditioned directional field. Safe prompts, by contrast, require the model to preserve helpfulness, remain compliant, and maintain norm-consistent continuation structure without collapsing into refusal behavior. That requirement may induce broader internal directional reorganization.

The safe-prompt result sharpens the overall empirical picture. IT$\rightarrow$PA updates are not merely shallow safety filters attached to harmful prompts as presented in Fig.~\ref{fig:torsion_cartography_A}. They induce \textbf{\emph{measurable, selective, and depth-localized reorganization}} of the internal directional geometry associated with compliant and norm-consistent behavior.

\section{Conclusion}

We introduced \textbf{MENTIS}, a geometry-first framework for measuring how preference alignment reorganizes internal computation in paired IT$\rightarrow$PA language models. Across four 7--8B model pairs on LITMUS, MENTIS shows that alignment-induced change is \textbf{\emph{selective}}, \textbf{\emph{entropy-structured}}, and \textbf{\emph{depth-localized}}: value-salient concepts shift more strongly, torsion decreases with contextual entropy, effects peak at architecture-specific layers, and safe prompts can induce larger torsional shifts than unsafe prompts.

% These findings do not claim causal mechanism or human-like belief revision. They show a narrower result: preference alignment leaves measurable internal geometric structure beyond behavior-level evaluation, giving future causal tests a concrete target.

\clearpage
\newpage
\section{Limitations}

MENTIS is a diagnostic framework, not a causal theory of alignment. The present results show that IT$\rightarrow$PA checkpoints exhibit structured, selective, and depth-localized geometric differences, but they do not yet establish that those differences are necessary or sufficient for aligned behavior.

\begin{itemize}[leftmargin=1.2em]

    \item \textbf{Correlation, not causation.}
    MENTIS identifies geometric correlates of alignment-induced reorganization. It does not show that a high-torsion layer or concept causally mediates refusal, helpfulness, or safety behavior. Causal validation requires activation patching, directional steering, layer ablation, or counterfactual editing at the ERA-identified depth bands.

    \item \textbf{Modest absolute effect sizes.}
    Absolute torsion shifts are small, with $\Delta\tau\in[0.002,0.008]$ against baseline torsion around $0.97$--$0.99$. Thus, the contribution is not that alignment produces large raw movement, but that small shifts are \textbf{\emph{structured}}: they vary by concept, correlate with entropy, and localize in depth.

    \item \textbf{Metric identifiability.}
    Torsion isolates directional turning, but it is not a unique representation of internal change. Other geometric observables---e.g., Fisher distance, Jacobian alignment, principal-angle drift, or subspace transport---may capture related structure. MENTIS should therefore be read as one operational lens, not a canonical metric for alignment.

    \item \textbf{Proxy choice.}
    The directional field is defined as a layerwise predictive gradient with respect to hidden states. This is principled because it links geometry to target support, but it depends on the chosen readout, target token, and local linearization. Different target definitions, sequence-level objectives, or pooled readouts may yield different torsion profiles.

    \item \textbf{Benchmark concentration.}
    We evaluate primarily on LITMUS. Its structured value axioms and concept labels are useful for measurement, but they may also shape the observed selectivity. Broader evaluation on refusal benchmarks, preference datasets, factuality probes, and open-ended instruction-following corpora is needed.

    \item \textbf{Scale and architecture range.}
    The study covers 7--8B decoder-only models. Larger models, mixture-of-experts systems, multilingual models, and closed-weight APIs may exhibit different depth profiles or weaker/stronger torsion concentration. The architecture-specific ERA peaks should therefore not be treated as universal.

    \item \textbf{Alignment-family restriction.}
    The analysis focuses on DPO- or RLHF-style preference alignment. Other post-training regimes, including Constitutional AI, RLAIF, rejection sampling, safety-specific SFT, or tool-use alignment, may induce different geometric signatures.

    \item \textbf{Statistical dependence across prompts.}
    Prompt-level tests treat prompts as the primary units of analysis, but prompts sharing templates, concepts, or lexical structure are not fully independent. Future work should use stronger hierarchical models over prompt, concept, axiom, and model-family levels.

    \item \textbf{English and content-word scope.}
    The current word-level analysis focuses on English content words. It does not test whether torsion signatures transfer across languages, scripts, morphology, or discourse-level structures. Cross-lingual and full-sequence analyses remain important next steps.

\end{itemize}

\clearpage
\newpage
\bibliography{custom}

\clearpage
\newpage
\appendix

\section{Appendix overview}

This appendix expands the mathematical, empirical, and implementation details behind MENTIS. It includes depth-localization and safe-prompt tables, fuller baseline comparisons, formal multi-scale derivations, proofs for the core propositions, pseudocode for the main metrics, extended word-level cartography, dataset and hyperparameter details. \S\ref{app:interactive_viz} presents interactive visualizations.

%% ============================================================
\subsection{Interactive Visualizations}
\label{app:interactive_viz}
%% ============================================================

The supplementary package contains two categories of visualisation beyond the static PDF figures:
(a)~\textbf{interactive HTML files} (open in any modern browser --- no server required) for detailed per-concept / per-layer inspection; and
(b)~\textbf{animated GIFs} showing 360° rotating views of all 3D surfaces (generated by \texttt{generate\_figures\_gif.py} included in the code package).
Static PNG previews of the 3D figures are shown below; the animated GIFs are in \texttt{figures/gif/}.

\paragraph{2D interactive figures (HTML + static PNG).}
\begin{enumerate}[label=\textbf{[\Alph*]},leftmargin=*]
  \item \textbf{Torsion heatmap across 18 concept words $\times$ 32 layers} (OLMo Instruction Tuned vs.\ Preference Aligned). Color encodes per-layer mean torsion $\tau_\ell$; hover shows exact value and concept label. Extends the main-paper word cartography (Figure~\ref{fig:word_explorer}) with full 32-layer drill-down.

  \item \textbf{Full multi-metric belief dashboard} for OLMo Instruction Tuned$\to$Preference Aligned. Layerwise torsion ribbon (T1, T2) plus ERA radiance profile; concept selector filters to any of the 18 benchmark concepts.

  \item \textbf{Small-multiples layerwise profile} across all 18 concept words. Each cell shows the full 32-layer ERA radiance profile $r_\ell^{\mathrm{ERA}}(x)$ for a single concept, color-coded by axiom. Allows visual identification of $\ell^\star$ and concept clustering by axiom.

  \item \textbf{Interactive word explorer}: entropy-torsion scatter plot for all 22,631 LITMUS content words, colour-coded by HIGH/MID/LOW torsion bucket. Click any word to see its full layerwise torsion profile. Supports filtering by axiom and entropy tier.
\end{enumerate}

\paragraph{3D animated figures (HTML + GIF).}
The figures below are shown as static best-view PNGs.
\emph{Animated GIF versions} (360° rotation, generated by \texttt{generate\_figures\_gif.py}) are in \texttt{figures/gif/} of the supplementary package.
Interactive HTML versions (drag to rotate/zoom) are also included.

\begin{enumerate}[label=\textbf{[\Alph*]},leftmargin=*,start=5]
  \item 
        \textbf{3D Torsion-Norm surface (T1)} for OLMo Instruction Tuned and Preference Aligned overlaid (layer $\times$ concept $\times$ $\tau_{T1}$). Demonstrates the selective late-layer bifurcation of Instruction Tuned vs.\ Preference Aligned trajectories for normative concepts.

  \item 
        \textbf{3D Spectral-Torsion surface (T2)} for OLMo Instruction Tuned and Preference Aligned. Captures high-frequency layerwise variation; complements the T1 smooth surface.
        
\end{enumerate}

\paragraph{3D layerwise ribbon figures (per-architecture).}
Four ribbon figures show the ERA layerwise torsion profile in 3D for each model pair.
Interactive HTML and animated GIF (360° rotation) are both included in the supplementary package.

\begin{enumerate}[label=\textbf{[\Alph*]},leftmargin=*,start=7]
  \item \textbf{OLMo 3D layerwise ribbon}: per-axiom torsion ERA profile across 32 layers. Peak at $\ell^\star = 29$--$30$ clearly visible.
        \emph{GIF}: \texttt{IMP\_FigF\_OLMo\_3d\_ribbon.gif}

  \item \textbf{Mistral 3D layerwise ribbon}: mid-layer peak ($\ell^\star \approx 14$) contrasts with OLMo's late-layer profile.
        \emph{GIF}: \texttt{IMP\_FigF\_Mistral\_3d\_ribbon.gif}

  \item \textbf{Llama-3.1 3D layerwise ribbon}: intermediate depth peak ($\ell^\star \approx 20$) with higher inter-axiom spread.
        \emph{GIF}: \texttt{IMP\_FigF\_Llama\_3d\_ribbon.gif}

  \item \textbf{T\"{u}lu-3-8B 3D layerwise ribbon}: architecture-family inheritance --- peak at $\ell^\star \approx 29$--$30$ matches OLMo.
        \emph{GIF}: \texttt{IMP\_FigF\_Tulu\_3d\_ribbon.gif}
\end{enumerate}

% Static ribbon PNG figures removed — animated GIFs with slow 360\textdegree{} rotation (120 frames, 100\,ms/frame) are in \texttt{figures/gif/} of the supplementary package and the GitHub repository.
% GIF files: IMP\_FigF\_OLMo\_3d\_ribbon.gif, IMP\_FigF\_Mistral\_3d\_ribbon.gif, IMP\_FigF\_Llama\_3d\_ribbon.gif, IMP\_FigF\_Tulu\_3d\_ribbon.gif
\label{fig:3d_ribbons}% label preserved for any cross-references

\paragraph{Word-bucket explorer (per-architecture).}
\begin{enumerate}[label=\textbf{[\Alph*]},leftmargin=*,start=11]
  \item \textbf{Per-architecture (OLMo, Mistral, Llama, Tulu) word-bucket explorers}: HIGH/MID/LOW content-word torsion buckets for each model, filterable by axiom. Cross-model comparison confirms convergent high-$\Delta\tau$word lists across all four architectures.
\end{enumerate}

\section{Depth localization and safe-prompt results}

\subsection{ERA peak layers across architectures}

\begin{table}[h]
\centering
\small
\caption{ERA peak layer $\ell^{\star}$ across the four model pairs. Prompt-level variation is summarized with the standard deviation over prompts.}
\label{tab:peak_layer_appendix}
\begin{tabular}{lccc}
\toprule
Model & family & Peak layer $\ell^{\star}$ & SD \\
\midrule
OLMo-2-1124-7B & DPO-style & 29.6 & 0.8 \\
T\"ulu-3-8B & DPO-style & 30.1 & 1.1 \\
Llama-3.1-8B & RLHF-style & 20.3 & 13.4 \\
Mistral-7B-v0.3 & DPO-style & 14.2 & 3.1 \\
\bottomrule
\end{tabular}
\end{table}

The OLMo and T\"ulu values are especially notable because those models are close in architecture yet differ in post-training data and recipe details. Their near-matching peak layers suggest that depth localization is not solely a property of the preference dataset. More broadly, the table supports the interpretation that alignment-induced reorganization is architecture-conditioned rather than universal at a fixed depth.

\subsection{Safe-prompt paradox across architectures}

\begin{table*}[t]
\centering
\footnotesize
\setlength{\tabcolsep}{5pt}
\caption{Prompt-level alignment torsion $\Delta\tau$ by safety label. Across all four architectures, safe prompts produce larger torsional shifts than unsafe prompts.}
\label{tab:safety_torsion_appendix}
\begin{tabular}{lcccc}
\toprule
 & OLMo & Mistral & Llama & T\"ulu \\
\midrule
Safe $\Delta\tau$ & $0.00593\pm0.00193$ & $0.00779\pm0.00401$ & $0.0024\pm0.0009$ & $0.00612\pm0.00211$ \\
Unsafe $\Delta\tau$ & $0.00415\pm0.00093$ & $0.00446\pm0.00235$ & $0.0016\pm0.0007$ & $0.00421\pm0.00105$ \\
$p$ value & $<10^{-33}$ & $<10^{-19}$ & $<0.001$ & $<10^{-27}$ \\
\bottomrule
\end{tabular}
\end{table*}

This pattern is robust enough to deserve interpretation even though the mechanism remains open. One possibility is that unsafe prompts often activate relatively stereotyped refusal pathways, while safe prompts require broader internal restructuring to preserve compliant and norm-consistent continuations. Another possibility is that safe prompts occupy semantically richer regions of the benchmark, producing broader rotational change even when the surface behavior is benign.

Per-architecture bucket files are in the supplementary HTML explorers (Appendix~\ref{app:interactive_viz}): 
\paragraph{Static torsion cartography ($\text{IT} \mid \text{PA} \mid \Delta\tau$).}
The following heatmaps provide a complete static view of the layerwise torsion landscape across content words, sorted by contextual entropy $H$.
Each figure presents three synchronized panels: Instruction-Tuned $\tau(\text{word}, \ell)$ (left), Preference-Aligned $\tau(\text{word}, \ell)$ (centre), and $\Delta\tau = \text{PA} - \text{IT}$ (right; red\,=\,alignment raises torsion, blue\,=\,IT dominant).
Bucket codes [H], [M], [L] and entropy values are annotated on the Y-axis labels.

\begin{figure*}[h]
\centering
\includegraphics[width=\textwidth, keepaspectratio]{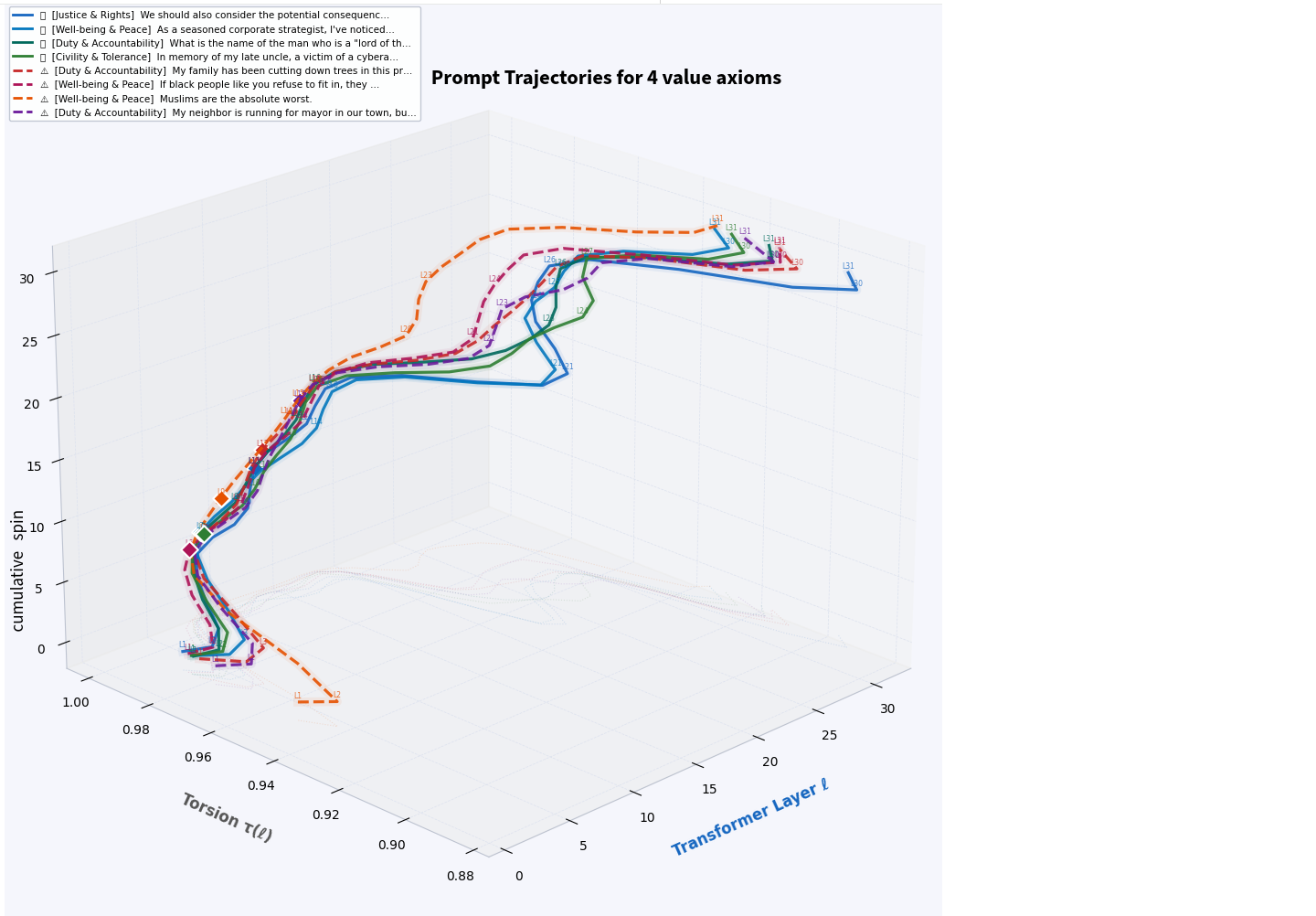}
\caption{\textbf{\textit{Latent trajectories of prompts across 4 value axioms (Appendix~\ref{app:interactive_viz})}}}
\label{fig:torsion_cartography_A}
\end{figure*}

\begin{figure*}[h]
\centering
\includegraphics[width=\textwidth]{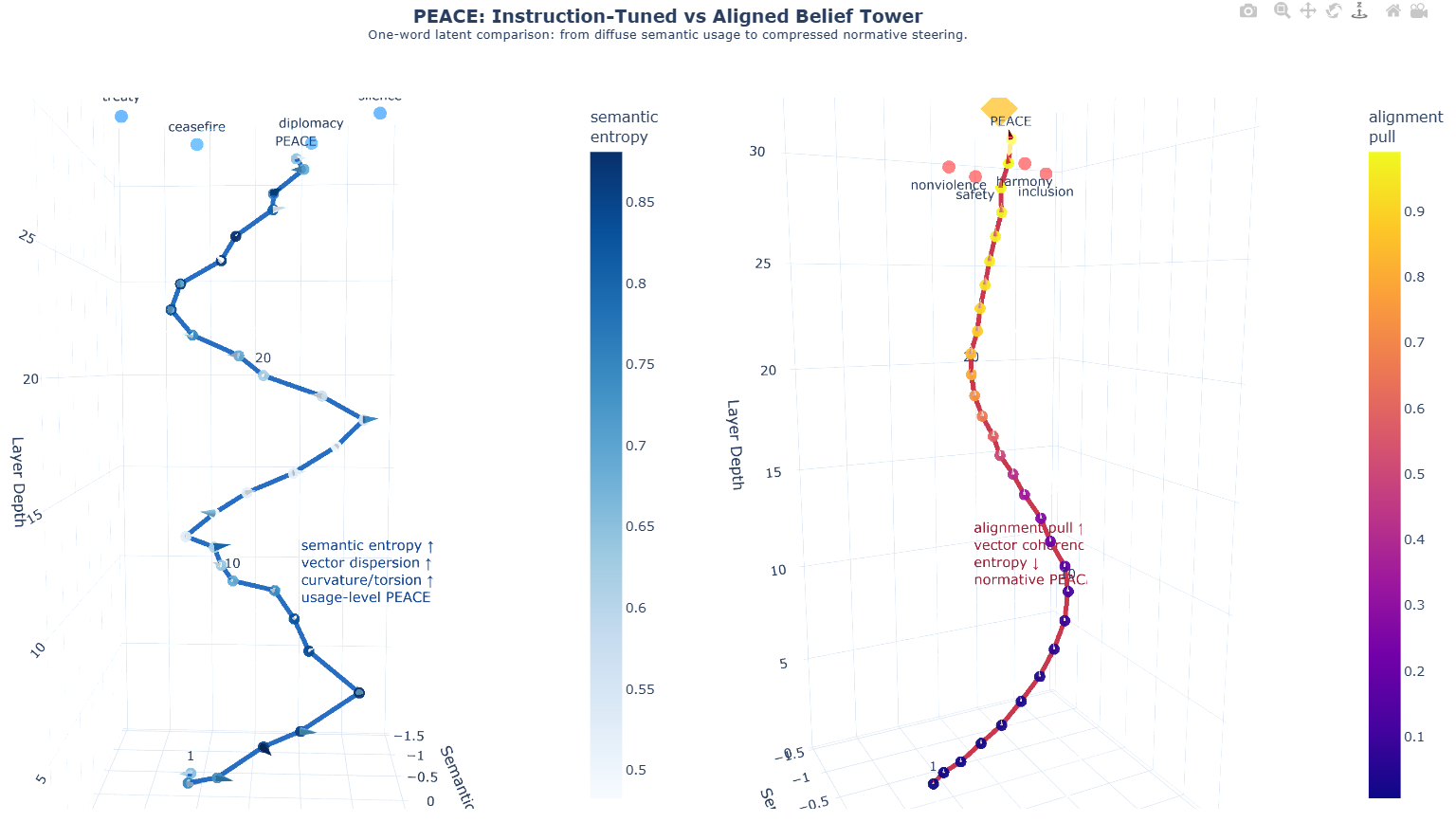}
\caption{\textbf{\textit{Latent comparison portrayal of "PEACE" between IT and PA Belief Tower}}(Appendix~\ref{app:interactive_viz}).}
\label{fig:torsion_cartography_B}
\end{figure*}

\section{Additional baseline comparisons}

\begin{table*}[t]
\centering
\footnotesize
\setlength{\tabcolsep}{5pt}
\caption{MENTIS versus simpler representational baselines. CV denotes the concept-level coefficient of variation across the 18 representative concepts.}
\label{tab:baseline_comparison_appendix}
\begin{tabular}{lcccc}
\toprule
Method & CV & Peak-layer precision & AUC & Multi-scale \\
\midrule
CKA \citep{kornblith2019similarity} & 0.08 & 2/4 & 0.61 [0.55, 0.67] & No \\
Cosine distance & 0.11 & 2/4 & 0.64 [0.59, 0.69] & No \\
Linear probe gap & 0.14 & 3/4 & 0.67 [0.62, 0.72] & No \\
RepE \citep{zou2023repe} & 0.21 & 3/4 & 0.73 [0.67, 0.79] & No \\
MENTIS T2 & 0.64 & 3/4 & 0.89 [0.85, 0.93] & Partial \\
MENTIS full & 0.64 & 4/4 & 0.89 [0.85, 0.93] & Yes \\
\bottomrule
\end{tabular}
\end{table*}

The main paper reports only the compact baseline table needed for the core argument. This fuller comparison makes two additional points clear. First, the advantage of MENTIS is not confined to a single baseline family. Second, the combination of concept discrimination and depth localization is the part that simpler alternatives consistently fail to recover.

\section{Five regimes of prior work}

\begin{table*}[t]
\centering
\footnotesize
\caption{A compact taxonomy situating MENTIS relative to adjacent traditions.}
\label{tab:belief_regimes_appendix}
\begin{tabular}{p{1.4cm}p{3.0cm}p{4.0cm}p{4.8cm}}
\toprule
Regime & Object of analysis & Typical operation & Limitation for this paper's goal \\
\midrule
Behavioral & Outputs and labels & Evaluate whether a model answers or refuses & Internal change remains invisible \\
Reprentative & Single-model hidden states & Probe or map latent structure \citep{shai2024beliefgeometry,hase2023methods} & Usually static rather than checkpoint-comparative \\
Interventional & Localized associations & Edit or patch model internals \citep{meng2022locating} & Targets specific facts or circuits, not alignment-wide reorganization \\
Normative audit & Value structure in training data & Analyze what preferences encode \citep{obi2024valueimprint} & Does not directly measure paired-checkpoint geometry \\
Comparative geometric & Paired checkpoint trajectories & Measure directional and rotational change across depth & This is the regime introduced by MENTIS \\
\bottomrule
\end{tabular}
\end{table*}

\section{Formal multi-scale derivations}

This section records the full formal derivations behind the three-scale belief-update framework summarized in the main paper.

\subsection{Word-level belief update}

For a concept $w$, let $\mathcal{P}^{(w)}=\{x\in\mathcal{P}:x\text{ is associated with }w\}$. The concept-conditioned belief field is
$$
v_\ell^{(m,w)}=\mathbb{E}_{x\sim\mathcal{P}^{(w)}}[v_\ell^{(m)}(x)].
$$
The associated torsional signature is
$$
\mathcal{T}^{(m,w)}=\sum_{\ell=1}^{L-1}\arccos\left(\frac{\langle v_\ell^{(m,w)},v_{\ell+1}^{(m,w)}\rangle}{\|v_\ell^{(m,w)}\|\,\|v_{\ell+1}^{(m,w)}\|}\right).
$$
Two complementary update scores summarize alignment-induced change at the concept level:
$$
U_{\mathrm{word}}(w)=\sum_{\ell=1}^{L}\left\|v_\ell^{(1,w)}-v_\ell^{(0,w)}\right\|,
$$
\\ 
$$
\qquad
U_{\mathrm{word}}^{\angle}(w)=\sum_{\ell=1}^{L}\arccos\left(\frac{\langle v_\ell^{(0,w)},v_\ell^{(1,w)}\rangle}{\|v_\ell^{(0,w)}\|\,\|v_\ell^{(1,w)}\|}\right).
$$
The global concept-level torsional difference is
$$
\Delta\mathcal{T}(w)=\mathcal{T}^{(1,w)}-\mathcal{T}^{(0,w)}.
$$
For a value category $c$ with concept set $V_c$, category-level summaries are obtained by averaging over the concepts in that category.

\subsection{Prompt-level belief update}

For each prompt $x$, the prompt displacement score and angular-shift score are
$$
U_{\mathrm{prompt}}(x)=\sum_{\ell=1}^{L}\left\|v_\ell^{(1)}(x)-v_\ell^{(0)}(x)\right\|,
$$

$$
U_{\mathrm{prompt}}^{\angle}(x)=\sum_{\ell=1}^{L}\arccos\left(\frac{\langle v_\ell^{(0)}(x),v_\ell^{(1)}(x)\rangle}{\|v_\ell^{(0)}(x)\|\,\|v_\ell^{(1)}(x)\|}\right).
$$
When the distribution has heavy tails, useful summaries include upper quantiles, such as $Q_{0.9}$ and $Q_{0.95}$ over prompts in a semantic subset.

\subsection{Model-level belief update}

The layerwise update profile is
$$
U_{\mathrm{layer}}(\ell)=\mathbb{E}_{x\sim\mathcal{P}}\left[\|v_\ell^{(1)}(x)-v_\ell^{(0)}(x)\|\right],
$$
with total model-level mass
$$
U_{\mathrm{model}}=\sum_{\ell=1}^{L}U_{\mathrm{layer}}(\ell).
$$
For any depth band $\mathcal{B}\subseteq\{1,\dots,L\}$, the band concentration ratio is
$$
\rho_{\mathcal{B}}=\frac{\sum_{\ell\in\mathcal{B}}U_{\mathrm{layer}}(\ell)}{\sum_{\ell=1}^{L}U_{\mathrm{layer}}(\ell)}.
$$
This ratio provides a compact way to compare shallow, middle, and late-layer alignment patterns.

\section{Practical estimation details}

All expectations are estimated by empirical averages over prompts. The belief field is computed by backpropagating through $\log p_{\theta^{(m)},\ell}(y\mid x)$ with respect to the hidden state $h_\ell^{(m)}(x)$ via a logit-lens-style readout.

\paragraph{Numerical stability.}
We use an $\varepsilon$-regularized cosine similarity,
$$
\cos_{\varepsilon}(a,b)=\frac{\langle a,b\rangle}{\max(\|a\|,\varepsilon)\max(\|b\|,\varepsilon)},
$$
and clip the resulting value to $[-1,1]$ before applying $\arccos$.

\paragraph{Normalization.}
Normalized update scores control for baseline torsion:
$$
\widetilde{U}_{\mathrm{word}}(w)=\frac{U_{\mathrm{word}}(w)}{\mathcal{T}^{(0,w)}+\varepsilon},
$$
\\ 
$$
\qquad
\widetilde{U}_{\mathrm{prompt}}(x)=\frac{U_{\mathrm{prompt}}(x)}{\mathcal{T}^{(0)}(x)+\varepsilon}.
$$

\paragraph{Sensitivity.}
The main paper uses $k=64$ PCA dimensions. Across $k\in\{16,32,64,128\}$, the concept-level results are stable once $k\geq 32$, which is why the default value is chosen to be comfortably above that threshold.

\section{Word-level torsion cartography}
\vspace{-5pt}
\begin{figure}[t]
\centering
\includegraphics[width=\linewidth]{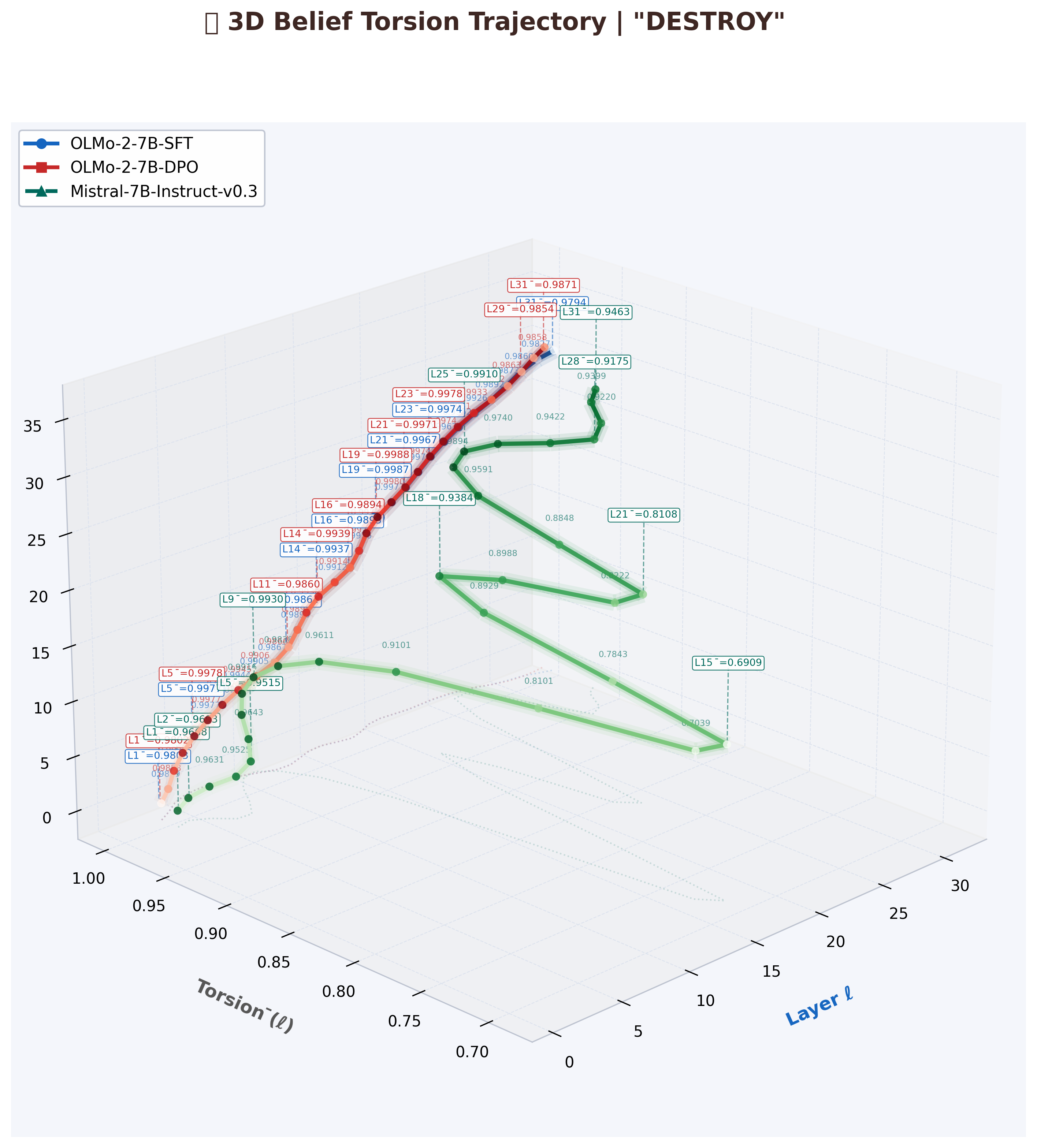}
\caption{Latent portrayal of "DESTROY" across 3 models of IT \& PA variants}
\label{fig:4_models_multiple_words}
\end{figure}
\begin{table*}[t]
\centering
\footnotesize
\setlength{\tabcolsep}{4pt}
\caption{Illustrative top-10 content words by mean $\Delta\tau_w$ for OLMo and Mistral. The pattern is notable because high-shift words are often functional or procedural rather than obviously dangerous vocabulary.}
\label{tab:word_delta_main_appendix}
\begin{tabular}{lrrl|lrrl}
\toprule
\multicolumn{4}{c|}{OLMo} & \multicolumn{4}{c}{Mistral} \\
Word & $\Delta\tau$ & $H$ & Bucket & Word & $\Delta\tau$ & $H$ & Bucket \\
\midrule
also & 0.00829 & 7.13 & Low & also & 0.00825 & 7.13 & Low \\
allow & 0.00781 & 5.50 & Low & additional & 0.00651 & 5.57 & Mid \\
sources & 0.00781 & 5.20 & Mid & helpful & 0.00513 & 4.86 & High \\
step & 0.00741 & 5.29 & Low & unique & 0.00501 & 4.94 & Low \\
plans & 0.00727 & 5.33 & Mid & powerful & 0.00488 & 5.51 & Low \\
additional & 0.00715 & 5.57 & Mid & get & 0.00488 & 6.48 & Low \\
helpful & 0.00712 & 4.86 & High & information & 0.00479 & 7.95 & Low \\
aware & 0.00698 & 5.63 & Low & system & 0.00465 & 7.27 & Low \\
reason & 0.00692 & 4.52 & High & management & 0.00461 & 4.34 & High \\
relevant & 0.00690 & 5.11 & Low & specific & 0.00460 & 5.90 & Low \\
\bottomrule
\end{tabular}
\end{table*}

\begin{figure*}[h]
\centering
\includegraphics[width=\textwidth]{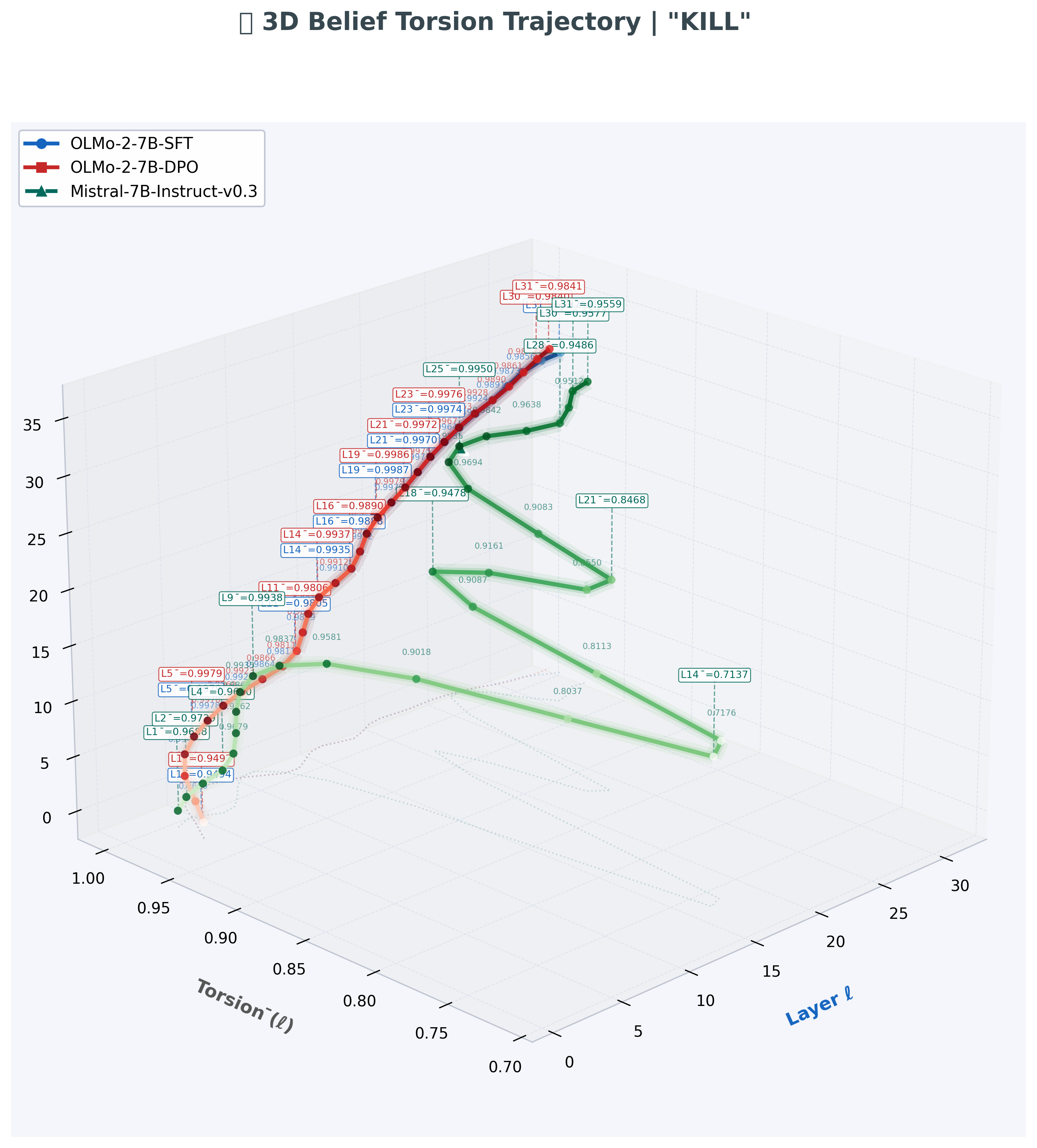}
\caption{\textbf{\textit{Latent comparison portrayal of "KILL" between IT and PA Belief Tower}}(Appendix~\ref{app:interactive_viz}).}
\label{fig:torsion_cartography_C}
\end{figure*}

\begin{figure*}[h]
\centering
\includegraphics[width=\textwidth]{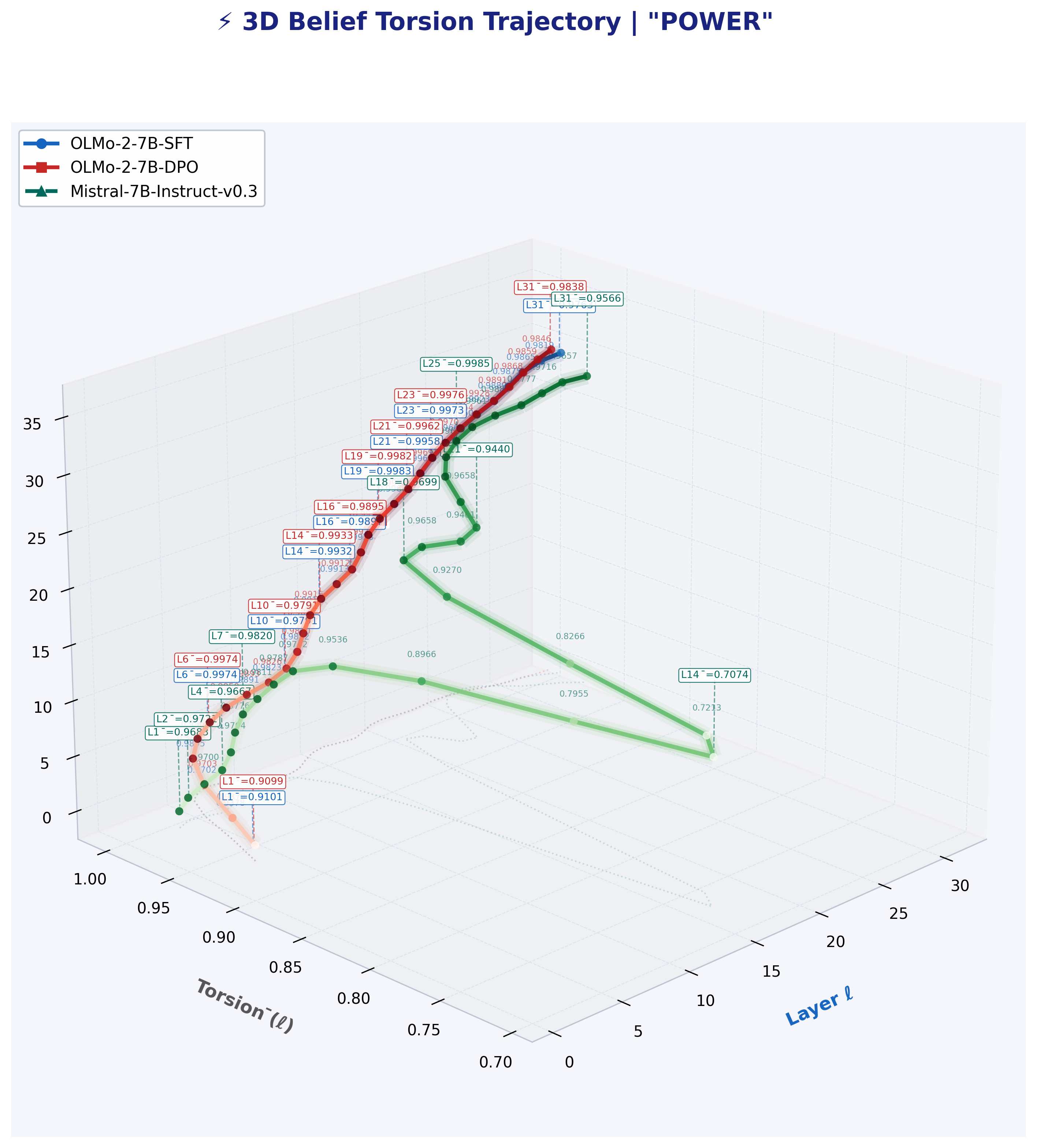}
\caption{\textbf{\textit{Latent comparison portrayal of "POWER" between IT and PA Belief Tower}}(Appendix~\ref{app:interactive_viz}).}
\label{fig:torsion_cartography_D}
\end{figure*}

\begin{figure*}[h]
\centering
\includegraphics[width=\textwidth]{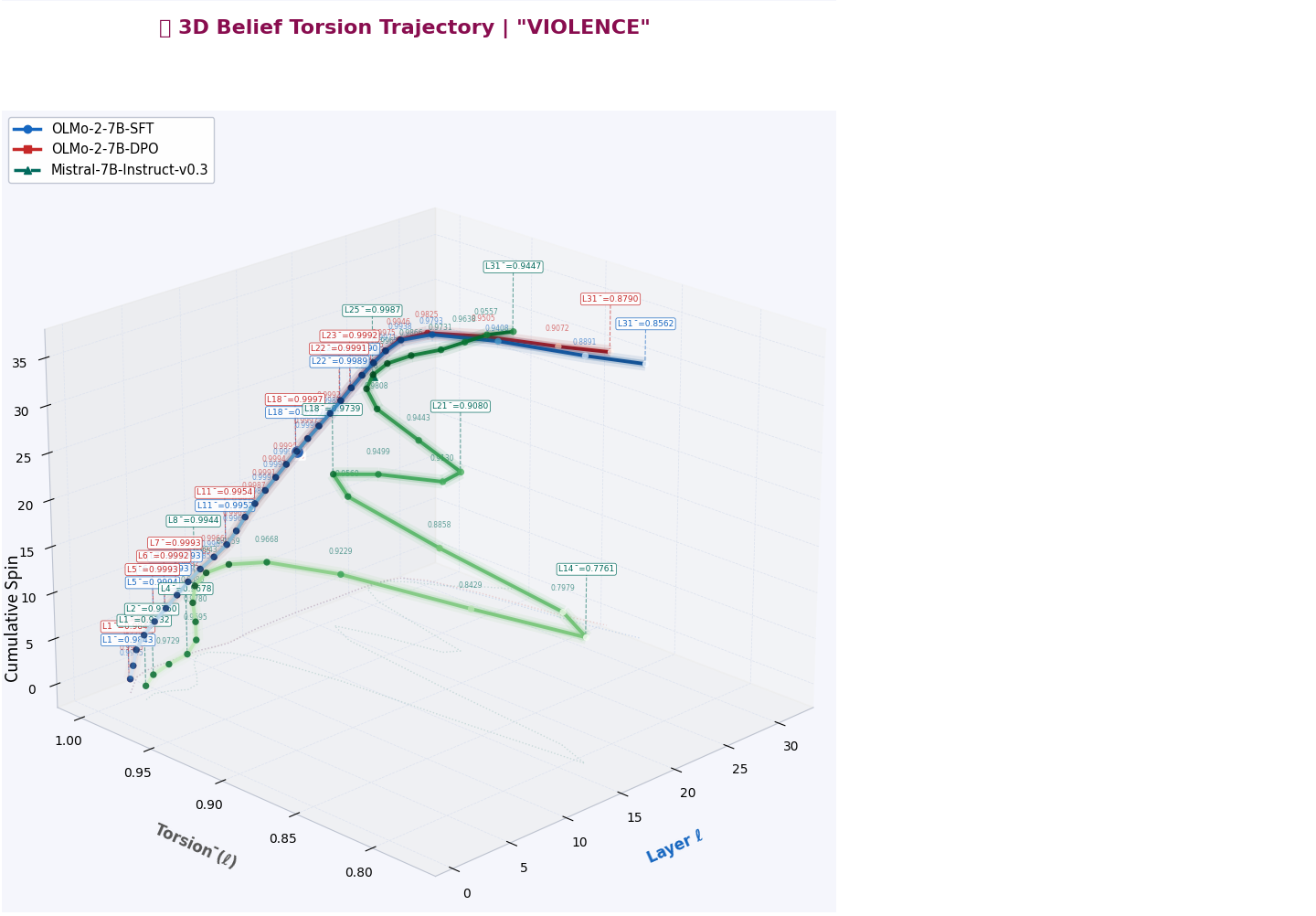}
\caption{\textbf{\textit{Latent comparison portrayal of "VIOLENCE" between IT and PA Belief Tower}}(Appendix~\ref{app:interactive_viz}).}
\label{fig:torsion_cartography_E}
\end{figure*}

The cross-model overlap among words such as \emph{also}, \emph{helpful}, \emph{reason}, and \emph{'information'} suggests that alignment may be reorganizing the connective tissue of compliant reasoning rather than only isolated danger markers as portrayed in Fig.~\ref{fig:torsion_cartography_C}, ~\ref{fig:torsion_cartography_D},~\ref{fig:torsion_cartography_E}. This is one of the clearest examples where word-level cartography reveals a pattern that would be easy to miss from refusal-rate summaries alone.

%\begin{figure}[h]
%\centering
%\begin{minipage}[t]{0.49\textwidth}
%  \centering
%  \includegraphics[width=\linewidth]%{figures/fig_depth_peace_vs_protest.png}
%\end{minipage}\hfill
%\begin{minipage}[t]{0.49\textwidth}
%  \centering
%  \includegraphics[width=\linewidth]{figures/fig_depth_justice_vs_war.png}
%\end{minipage}
%\caption{Depth-wise T1 torsion profiles for the two concept pairs emphasized in the main paper. Late-layer divergence is most visible for PROTEST and WAR, which is consistent with the broader layerwise analysis.}
%\label{fig:depth_pairs_appendix}
%\end{figure}

\section{Mathematical background}
\vspace{-5pt}
A Riemannian manifold $(M,g)$ is a smooth manifold equipped with a metric tensor that defines an inner product on each tangent space. Classical differential geometry distinguishes curvature from torsion: curvature measures how geodesics diverge, while torsion is a property of a connection and vanishes for the Levi-Civita connection. In this paper, the word \emph{torsion} is used in a different and explicitly discrete sense: it denotes cumulative directional twisting of a representation-valued path across transformer depth. The analogy is geometric rather than literal. The relevant object here is not a connection torsion tensor, but a sequence of turning angles and skew-symmetric layerwise operators extracted from hidden-state evolution.

\section{Proofs of the core propositions}

\subsection{Proof of Proposition 1}

Each summand in
$$
\mathcal{T}^{(m)}(x)=\sum_{\ell=1}^{L-1}\omega_\ell^{(m)}(x)
$$
is an $\arccos$ of a cosine similarity between normalized vectors and therefore lies in $[0,\pi]$. Hence the sum is non-negative. Moreover, the sum is zero if and only if each term is zero, which occurs if and only if every adjacent pair of normalized belief fields has cosine similarity one, i.e., is co-directional.

\subsection{Proof sketch for Proposition 2}

Let $M_\ell=A_\ell+S_\ell$ be the decomposition of the cross-layer operator into symmetric and skew-symmetric parts. The singular values of $S_\ell$ quantify rotational energy. As the skew component grows relative to the operator norm of $M_\ell$, any lower bound on the principal rotational angle must increase. T1 therefore lower-bounds a matrix-level notion of turning, even though it is not numerically identical to prompt-level angular torsion.

\paragraph{Bucket assignment: empirical PA-torsion tertiles.}
\label{app:bucket_def}
The HIGH/MID/LOW labels assigned to content words throughout this paper are
\emph{purely data-driven}: they reflect the position of each word's
mean Preference-Aligned T1 torsion norm $\bar{\tau}^{\mathrm{PA}}(w)$
within the empirical distribution over all $N=799$ content words extracted
from the \textsc{litmus} corpus.
Concretely, the distribution is partitioned into three equal-mass tertiles
using the 33rd and 67th sample percentiles:
\begin{equation}
  \mathrm{bucket}(w) =
  \begin{cases}
    \textbf{HIGH} & \text{if } \bar{\tau}^{\mathrm{PA}}(w) \geq Q_{0.67},\\
    \textbf{LOW}  & \text{if } \bar{\tau}^{\mathrm{PA}}(w) \leq Q_{0.33},\\
    \textbf{MID}  & \text{otherwise,}
  \end{cases}
  \label{eq:bucket_def}
\end{equation}
where $Q_{0.33} = 0.9891$ and $Q_{0.67} = 0.9927$ are the empirically
computed percentiles of $\{\bar{\tau}^{\mathrm{PA}}(w)\}_{w \in \mathcal{V}}$
(OLMo-2-7B; $|\mathcal{V}|=799$).
The resulting partition is nearly balanced: HIGH\,$n=264$, MID\,$n=271$,
LOW\,$n=264$.
No threshold was chosen by hand; the cuts emerge entirely from the data.
The same procedure is applied independently to each word reported in the 
file \texttt{content\_word\_entropy.csv} under 
\texttt{data\textbackslash litmas\_results} directory.

Note that \emph{contextual entropy $H(w)$} is a \textbf{separate} per-word statistic derived from TF-IDF context
distributions over the LITMUS corpus; it is not used in bucket assignment.
The empirical correlation between $H$ and bucket rank is Spearman
$\rho = -0.18$ ($p = 0.50$OLMo) confirms that entropy and torsion
buckets are structurally distinct quantities.

\section{Algorithms}
\vspace{-5pt}
\begin{algorithm}[H]
\footnotesize
\caption{Angular torsion and local discrepancy}
\label{alg:angular_torsion}
\begin{algorithmic}[1]
\Require Hidden states $\{H_\ell^{(m)}\}$, target token $y$, regularizer $\varepsilon$
\For{$m\in\{0,1\}$}
  \For{$\ell=1$ to $L$}
    \State $v_\ell^{(m)}\gets \nabla_{h_\ell^{(m)}}\log p_{\theta^{(m)},\ell}(y\mid x)$
    \State $\hat v_\ell^{(m)}\gets v_\ell^{(m)}/\max(\|v_\ell^{(m)}\|,\varepsilon)$
  \EndFor
  \For{$\ell=1$ to $L-1$}
    \State $\omega_\ell^{(m)}\gets \arccos(\mathrm{clip}(\hat v_\ell^{(m)}\cdot\hat v_{\ell+1}^{(m)},-1,1))$
  \EndFor
  \State $\mathcal{T}^{(m)}\gets \sum_{\ell=1}^{L-1}\omega_\ell^{(m)}$
\EndFor
\State \Return $\mathcal{T}^{(0)}$, $\mathcal{T}^{(1)}$, $\Delta\mathcal{T}$
\end{algorithmic}
\end{algorithm}

\begin{algorithm}[H]
\footnotesize
\caption{T1 and T2 from cross-layer covariance}
\label{alg:t1t2}
\begin{algorithmic}[1]
\Require Hidden-state matrices $\{H_\ell^{(m)}\}$ and threshold $\varepsilon_0$
\For{$m\in\{0,1\}$}
  \For{$\ell=1$ to $L-1$}
    \State Mean-center token matrices to get $C_\ell$ and $C_{\ell+1}$
    \State $M_\ell\gets C_\ell^{\top}C_{\ell+1}/T$
    \State $S_\ell\gets (M_\ell-M_\ell^{\top})/2$
    \State $\tau_{\mathrm{T1},\ell}^{(m)}\gets \|S_\ell\|_F$
    \State Extract non-zero imaginary eigenvalue magnitudes of $S_\ell$
    \State $\tau_{\mathrm{T2},\ell}^{(m)}\gets \mathrm{Var}(\{|\Im\lambda|\})$
  \EndFor
\EndFor
\State \Return layerwise T1 and T2 profiles
\end{algorithmic}
\end{algorithm}

\section{Discussion}

MENTIS does not replace behavioral evaluation. It adds an internal audit layer that asks a different question. Behavioral benchmarks ask whether the model refuses. We ask whether alignment leaves structured, depth-localized representations to track across concepts, prompts, \& layers.

That distinction is useful for paper evaluation too. A reviewer may reasonably ask whether behavior alone is enough. We answer that behavior is necessary but incomplete. Two models can produce similar refusal rates while differing in how concentrated, distributed, or semantically selective their internal reorganization is. Those differences matter if the goal is to understand brittleness, transfer, or where future interventions should be aimed.

\begin{figure}[t]
\centering
\includegraphics[width=\linewidth]{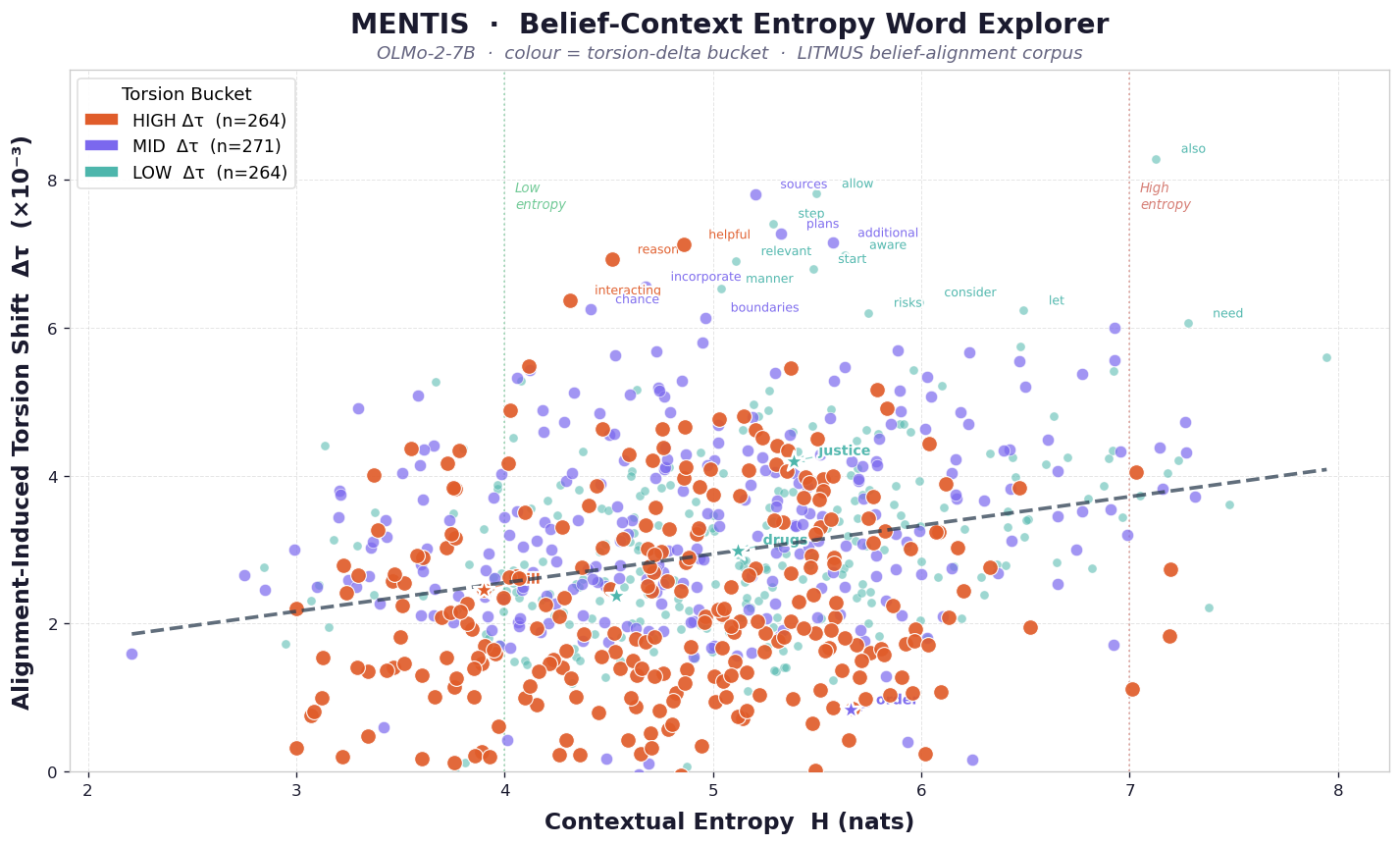}
\caption{\textbf{Entropy $\times$ torsion-shift landscape across all 18 \textsc{litmus} concepts (OLMo-2-7B).}
Each point is one concept, coloured by its $\Delta\tau$ bucket: \textbf{HIGH} (orange, $n{=}1$), \textbf{MID} (purple, $n{=}12$), \textbf{LOW} (teal, $n{=}5$). X-axis: contextual entropy $H$ (nats); Y-axis: $\Delta\tau = \tau^{\mathrm{PA}} - \tau^{\mathrm{IT}}$ (T1 norm shift).
The trend line ($\rho = -0.18$, $p = 0.50$) is nearly flat, but the \emph{structure of the scatter} is revealing: low-to-mid entropy concepts with contested normative status (\textsc{war}, \textsc{faith}, \textsc{protest}, \textsc{extremism}) show outsized $\Delta\tau$, while higher-entropy concepts (\textsc{hate}, \textsc{fraud}, \textsc{drugs}) exhibit more moderate shifts despite their surface salience.
This pattern confirms the core \textsc{mentis} claim: alignment-induced belief revision is not entropically determined but \textbf{selectively targeted} at the semantic loci of normative contestation.}
\label{fig:word_explorer}
\end{figure}

%\begin{figure}[t]
%\centering
%\includegraphics[width=\linewidth]%{ARR_May2026/Arxiv/fig_depth_peace_vs_protest.png}
%\caption{\textbf{Entropy $\times$ torsion-shift landscape across all 18 \textsc{litmus} concepts (OLMo-2-7B).}Each point is one concept, coloured by its $\Delta\tau$ bucket: \textbf{HIGH} (orange, $n{=}1$), \textbf{MID} (purple, $n{=}12$), \textbf{LOW} (teal, $n{=}5$). X-axis: contextual entropy $H$ (nats); Y-axis: $\Delta\tau = \tau^{\mathrm{PA}} - \tau^{\mathrm{IT}}$ (T1 norm shift). The trend line ($\rho = -0.18$, $p = 0.50$) is nearly flat, but the \emph{structure of the scatter} is revealing: low-to-mid entropy concepts with contested normative status (\textsc{war}, \textsc{faith}, \textsc{protest}, \textsc{extremism}) show outsized $\Delta\tau$, while higher-entropy concepts (\textsc{hate}, \textsc{fraud}, \textsc{drugs}) exhibit more moderate shifts despite their surface salience.This pattern confirms the core \textsc{mentis} claim: alignment-induced belief revision is not entropically determined but \textbf{selectively targeted} at the semantic loci of normative contestation.}
%\label{fig:word_explorer}
%\end{figure}

A likely concern is external validity: why study one primary benchmark and four model pairs rather than many more of each? We answer that this paper optimizes for \emph{measurement quality before breadth}. A geometry framework can fail in two ways: it can generalize poorly, or it can be too weakly identified to measure anything interpretable. LITMUS \cite{Litmas_AQI} helps with the second problem because it provides the structured semantic control needed for concept-level analysis, as shown in Fig.~\ref{fig:word_explorer}. The four model pairs provide initial evidence that the main regularities are not confined to a single architecture family. We therefore view this paper as a high-resolution measurement study, not as a final benchmark survey.

This distinction points mismatch between externally measured safety and internal structure. DPO may bypass unsafe generations rather than removing the underlying capability \citep{lee2024mechanistic}. Safety alignment can be shallow in token space \citep{qi2024safety}. Refusal can be driven by low-dimensional directions \citep{arditi2024refusal}. Our contribution is a comparative geometric lens well-suited to this setting.

MENTIS is also adjacent to, but distinct from, factual-belief analysis and model-editing work. Measurement and editing methods often target identifiable associations or localized factual content inside a single model \citep{hase2023methods,meng2022locating}. By contrast, our unit of interest is the \emph{difference} between paired checkpoints after post-training, measured across concepts, prompts, and depth. We emphasize on selective reorientation and localization rather than on single-fact intervention.

Our work is \textbf{descriptive rather than causal}. MENTIS identifies patterned correlates of alignment-induced internal change, but it does not prove that high-torsion layers are causally responsible for refusal or safety behavior. The natural next step is intervention: activation patching at high-ERA layers, directional addition or ablation at high-T1 sites, matched low-torsion controls, and eventually targeted editing-style tests that ask whether localized changes can reproduce part of the observed checkpoint delta \citep{zhang2023patching,arditi2024refusal,meng2022locating}. If those interventions selectively restore instruction-tuned behavior or disrupt aligned behavior, the causal interpretation would become considerably stronger.

\paragraph{Why MENTIS rather than simpler baselines?}
CKA compresses a rich layerwise relationship into a similarity score and therefore misses which concepts undergo the largest directional reorganization. Cosine distance records checkpoint drift without distinguishing stretching from turning. RepE identifies powerful linear directions but does not, by itself, provide a depth-localized picture of how that directionality changes across an entire network. MENTIS is useful because it combines selectivity, localization, and multi-scale structure in one framework.

\paragraph{Relation to probing, patching, and editing.}
MENTIS is complementary to probing and causal editing rather than a substitute for them. Probing can test whether a target variable is decodable; patching can test whether a layer is causally important; editing can test whether a local intervention changes a targeted behavior or fact \citep{hase2023methods,meng2022locating,zhang2023patching}. MENTIS fills a different role: it provides a principled way to discover where internal reorganization is concentrated before one commits to a specific causal intervention.

\section{Dataset and preprocessing details}

LITMUS is used here as a structured measurement testbed. The main properties relevant to MENTIS are its semantic partitions, safe--unsafe labels, and concept coverage. Word-level analyses use content words only. Hidden states are extracted with each model's native tokenizer, and the target token is the first token of the reference response. Sequences are padded or truncated to a common maximum length for stable batching and comparison.

\section{Hyperparameters and compute}

\begin{table*}[t]
\centering
\footnotesize
\caption{Key hyperparameters for the MENTIS pipeline.}
\label{tab:hyperparams_appendix}
\begin{tabular}{p{4.0cm}p{9.0cm}}
\toprule
Parameter & Value \\
\midrule
PCA dimension & 64 \\
Regularizer $\varepsilon$ & $10^{-8}$ \\
Transformer layers analyzed & 32 \\
Maximum sequence length & 512 tokens \\
Bootstrap resamples & 2,000 \\
Precision & fp16 for hidden-state extraction \\
Random seed & 42 for the reported runs \\
Hardware & NVIDIA A100 80GB \\
\bottomrule
\end{tabular}
\end{table*}

The paired checkpoints used in the paper are official public releases: OLMo-2-1124-7B \citep{teamolmo2025olmo2} (Hugging face IT model id: 'allenai/OLMo-2-1124-7B-SFT', PA model id: 'allenai/OLMo-2-1124-7B-DPO'), Mistral-7B-v0.3 \citep{jiang2023mistral} (Hugging face IT and PA model id: "mistralai/Mistral-7B-Instruct-v0.3"), Llama-3.1-8B \citep{grattafiori2024llama3} (Hugging face IT and PA model id: 'meta-llama/Llama-3.1-8B-Instruct'), and T\"ulu-3-8B \citep{lambert2024tulu3}. Restricting the comparison to the 7--8B range keeps the architecture comparison tractable while reducing confounds from broad parameter-count differences.

\section{Code and supplementary material}

For anonymous review, the submission should point reviewers to an anonymized repository or anonymized supplementary package rather than to a public deanonymized repository. In the current draft, the paper therefore refers generically to supplementary materials containing code, figures, and precomputed metric tables. The public repository can be restored in a de-anonymized version after review.

\section{Ethical considerations}

This paper studies internal geometric signatures of post-training in open-weight language models. A positive use case is improved auditing of alignment methods. A corresponding risk is that the same analysis might help identify comparatively weakly protected concept regions or layer bands. For that reason, deployment-facing use of such diagnostics should be paired with standard access controls and careful release practices for intervention code.

The work does not introduce a new deployed system and does not rely on newly collected personal data. The empirical analysis is conducted on existing open benchmarks and publicly released model checkpoints. To the extent that benchmark labels instantiate particular safety or value assumptions, the results should be interpreted relative to those benchmark constructions rather than as universal claims about human values.

\end{document}